\documentclass[lettersize,journal]{IEEEtran}
\usepackage{amsmath,amsfonts}
\usepackage{algorithmic}
\usepackage{array}
\usepackage{textcomp}
\usepackage{stfloats}
\usepackage{url}
\usepackage{verbatim}
\usepackage{graphicx}
\def\BibTeX{{\rm B\kern-.05em{\sc i\kern-.025em b}\kern-.08em
    T\kern-.1667em\lower.7ex\hbox{E}\kern-.125emX}}
\usepackage{balance}

\usepackage{times}
\usepackage{epsfig}
\usepackage{graphicx}
\usepackage{amsmath}
\usepackage{amssymb}
\usepackage[accsupp]{axessibility} 
\usepackage{pifont}
\usepackage{wrapfig}
\usepackage{diagbox}
\usepackage{bm}
\usepackage{adjustbox}
\usepackage{booktabs}

\usepackage{cases}
\usepackage{graphicx}
\usepackage{graphicx}
\usepackage{adjustbox}
\usepackage{booktabs} 
\usepackage{amsmath}
\usepackage{algorithm}
\usepackage{algorithmic} 
\usepackage{multirow}
\usepackage{enumitem}
\usepackage[thicklines]{cancel}
\usepackage{textcomp,booktabs}
\usepackage{colortbl}
\usepackage{pifont}
\usepackage[numbers]{natbib}

\definecolor{mygray}{gray}{.95}
\definecolor{CColor}{rgb}{0.01,0.31,0.59}
\definecolor{GGray}{rgb}{0.80,0.90,1}
\definecolor{Shady}{rgb}{0.9,0.9,0.9}
\definecolor{kaistblue}{RGB}{20,135,200}
\definecolor{kaistdarkblue}{RGB}{0,65,145}
\definecolor{urbanablue}{RGB}{19,41,75}
\definecolor{urbanaorange}{RGB}{232,74,39}
\definecolor{drp}{rgb}{0.53,0.15,0.34}
\usepackage[breaklinks=true,letterpaper=true,colorlinks,bookmarks=false]{hyperref}

\usepackage[table]{xcolor}

\usepackage{microtype}
\usepackage{booktabs}

\usepackage{amsmath}
\usepackage{amssymb}
\usepackage{mathtools}
\usepackage{amsthm}

\theoremstyle{plain}
\newtheorem{theorem}{Theorem}[section]

\theoremstyle{definition}
\newtheorem{definition}[theorem]{Definition}

\theoremstyle{remark}

\usepackage[textsize=tiny]{todonotes}
\usepackage{algorithm}
\usepackage{multirow}
\usepackage{enumitem}
\usepackage{placeins}
\newcommand{\copyed}[1]{\textcolor{black}{#1}} 

\definecolor{mygray}{gray}{.9}
\definecolor{CColor}{rgb}{0.01,0.31,0.59}
\definecolor{GGray}{rgb}{0.80,0.90,1}
\definecolor{Shady}{rgb}{0.9,0.9,0.9}
\definecolor{kaistblue}{RGB}{20,135,200}
\definecolor{kaistdarkblue}{RGB}{0,65,145}
\definecolor{urbanablue}{RGB}{19,41,75}
\definecolor{urbanaorange}{RGB}{232,74,39}
\definecolor{drp}{rgb}{0.53,0.15,0.34}

\DeclareMathOperator*{\argmin}{arg\,min}

\usepackage{graphicx}
\usepackage{microtype}
\usepackage{booktabs} 
\usepackage{graphicx}

\usepackage[misc]{ifsym}

\usepackage{titletoc}
\usepackage{tocloft}
\begin{document}
\title{Efficient and Effective Weight-Ensembling Mixture of Experts for Multi-Task Model Merging}
\author{
Li Shen$^{*}$,
Anke Tang$^{*}$, 
Enneng Yang$^{\dagger}$, 
Guibing Guo, 
Yong Luo$^{\dagger}$,
Lefei Zhang, 
Xiaochun Cao, \\
Bo Du,
and Dacheng Tao,~\IEEEmembership{Fellow,~IEEE}
\thanks{
$^{*}$Li Shen and Anke Tang contributed equally to this work. $^{\dagger}$Enneng Yang and Yong Luo are the corresponding authors.

This paper is an extended version of our previous paper on ICML 2024~\cite{tang2024merging}. In this version, we propose an Efficient Weight-Ensembling Mixture of Experts model (E-WEMoE), that significantly reduces both the number of trainable parameters and the total parameters compared to the previous version.

Li Shen and Xiaochun Cao are with Sun Yat-sen University, China. E-mail: mathshenli@gmail.com, caoxiaochun@mail.sysu.edu.cn.

Anke Tang, Yong Luo, Lefei Zhang, and Bo Du are with Wuhan University, China. E-mail: anketang@whu.edu.cn, yluo180@gmail.com, \{zhanglefei, dubo\}@whu.edu.cn.

Enneng Yang and Guibing Guo are with Northeastern University, China. E-mail: ennengyang@stumail.neu.edu.cn, guogb@swc.neu.edu.cn.

Dacheng Tao is with Nanyang Technological University, Singapore. E-mail: dacheng.tao@gmail.com.
}}


\maketitle

\begin{abstract}
  Multi-task learning (MTL) leverages a shared model to accomplish multiple tasks and facilitate knowledge transfer. Recent research on task arithmetic-based MTL demonstrates that merging the parameters of independently fine-tuned models can effectively achieve MTL. However, existing merging methods primarily seek a static optimal solution within the original model parameter space, which often results in performance degradation due to the inherent diversity among tasks and potential interferences. To address this challenge, in this paper, we propose a Weight-Ensembling Mixture of Experts (WEMoE) method for multi-task model merging. Specifically, we first identify critical (or sensitive) modules by analyzing parameter variations in core modules of Transformer-based models before and after fine-tuning. Then, our WEMoE statically merges non-critical modules while transforming critical modules into a mixture-of-experts (MoE) structure. During inference, expert modules in the MoE are dynamically merged based on input samples, enabling a more flexible and adaptive merging approach. Building on WEMoE, we further introduce an efficient-and-effective WEMoE (E-WEMoE) method, whose core mechanism involves eliminating non-essential elements in the critical modules of WEMoE and implementing shared routing across multiple MoE modules, thereby significantly reducing both the trainable parameters, the overall parameter count, and computational overhead of the merged model by WEMoE. Experimental results across various architectures and tasks demonstrate that both WEMoE and E-WEMoE outperform state-of-the-art (SOTA) model merging methods in terms of MTL performance, generalization, and robustness.
\end{abstract}

\begin{IEEEkeywords}
Ensembling Learning, Multi-task Learning, Mixture of Experts, Model Fusion 
\end{IEEEkeywords}

\section{Introduction}
\label{section:introduction}

The goal of multi-task learning (MTL) is to utilize a single model to perform multiple related tasks concurrently, thereby facilitating information sharing and knowledge transfer among the tasks. In recent years, the rapid development of deep learning has prompted a learning paradigm shift, where the mainstream paradigm now focuses on fine-tuning downstream tasks using powerful pre-trained models, rather than training an expert model from scratch~\citep{radfordLanguageModelsAre2019,heMaskedAutoencodersAre2021,DBLP:journals/ijautcomp/WangCQGWWTG23,chungScalingInstructionFinetunedLanguage2022,zhengLearnModelFineTuning2023,DBLP:journals/ijautcomp/CaoLHS24}. This shift typically results in significant reductions in both data requirements and computational resources. Additionally, the open-source ethos of the deep learning community has encouraged developers to release a vast array of expert models fine-tuned on various downstream tasks. To date, over one million diverse models have been made available on Hugging Face~\footnote{\url{https://huggingface.co/models}}. These above diverse factors have given rise to a new MTL paradigm, enabling the direct merging of multiple independently trained expert models to create a multi-task model without requiring access to their original training data~\citep{liDeepModelFusion2023,DBLP:journals/ijautcomp/LinGGZZL23,yadav2024survey,Survery_ModelMerging_2024}. 

However, due to potential task conflicts and interferences among multiple tasks, simply merging parameters from independently fine-tuned models may lead to a sharp decline in MTL performance~\cite{izmailovAveragingWeightsLeads2019,wortsmanModelSoupsAveraging2022}. Recently, an increasing number of studies have aimed to address the MTL performance degradation resulting from model merging~\cite{Survery_ModelMerging_2024}. A notable example is task arithmetic~\cite{ilharcoEditingModelsTask2023}, which introduces the concept of `task vectors' to extract task-specific knowledge from the fine-tuned models. By linearly weighting the task-private knowledge of multiple tasks into the pre-trained model, task arithmetic enhances the model's ability to process multiple downstream tasks. Inspired by task arithmetic, recent advancements have proposed techniques to alleviate sign conflicts among task vectors~\cite{yadavResolvingInterferenceWhen2023,guodong24neurips}, merge them in a fine-grained manner~\cite{matenaMergingModelsFisherWeighted2022,jinDatalessKnowledgeFusion2023,yangAdaMergingAdaptiveModel2023}, or combine task vectors within subspaces~\cite{tangConcreteSubspaceLearning2023,yu2024language,wang2024localizing,huang2024emr}. While these methods have considerably improved task arithmetic's performance, a noticeable performance gap still exists between the merged MTL model and the independently fine-tuned expert model (or the joint-trained MTL model).

\begin{figure*}[t]
    \centering
    \begin{minipage}{0.3\textwidth}
        \centering
        \includegraphics[width=\linewidth]{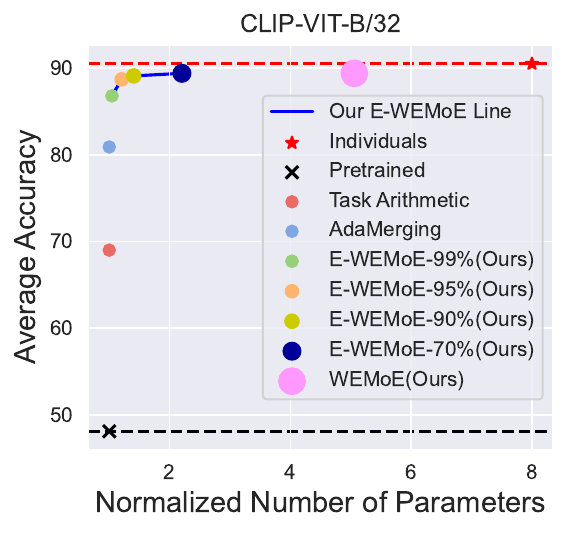}
    \end{minipage}
    \begin{minipage}{0.3\textwidth}
        \centering
        \includegraphics[width=\linewidth]{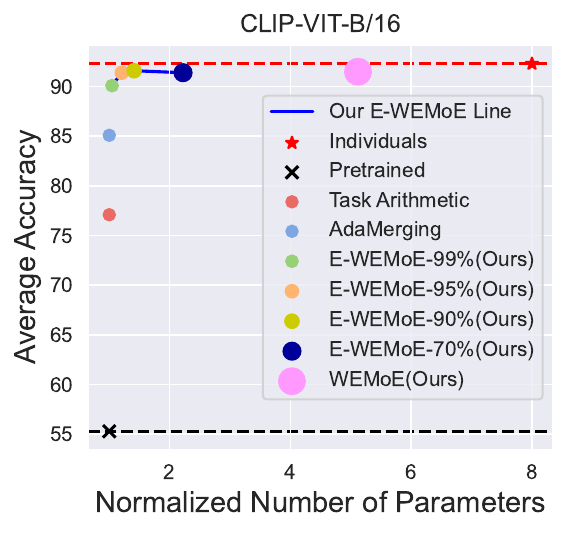}
    \end{minipage}
    \begin{minipage}{0.3\textwidth}
        \centering
        \includegraphics[width=\linewidth]{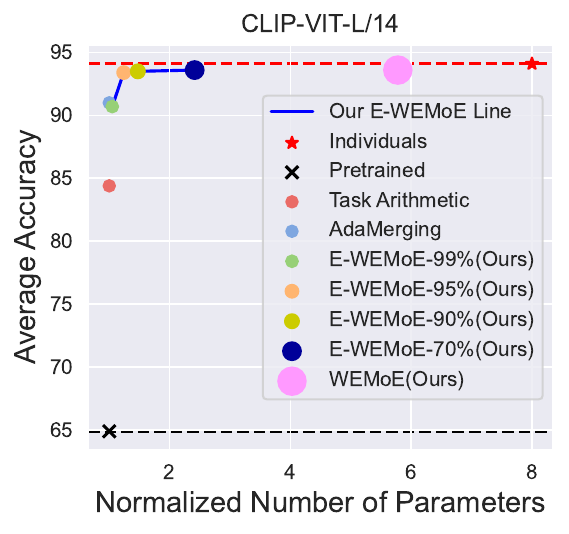}
    \end{minipage}
    \caption{
    Comparison of the relationship between parameter count and performance across various model merging methods. 
    }
    \label{fig:scatter}
\vskip -0.15in
\end{figure*}

In this paper, we claim that the observed performance gap primarily arises from the fact that existing model merging methods focus on finding a \textit{static} multi-task optimal solution within the original parameter space. This approach limits the effectiveness of merging models for different instances, as the inherent diversity among samples implies that the optimal model combination is often task- or instance-specific. Therefore, an ideal merging model should adapt \textit{dynamically} according to the particular task or instance being processed.

To address this challenge, we propose a novel approach, \textbf{W}eight-\textbf{E}nsembling \textbf{M}ixture \textbf{o}f \textbf{E}xperts (WEMoE), to dynamically merge Vision Transformers (ViT)-based models. More specifically, we first analyze the core modules--Attention (Att) and Multilayer Perceptron (MLP)--in multiple ViTs and observe that, compared to the pre-trained model, the MLP module exhibits more significant parameter changes after fine-tuning on downstream tasks, whereas the Att module shows less variation. This indicates that the MLP parameters capture more task-specific information, making them more susceptible to task conflicts during the merging process. Based on these findings, we adopt a traditional static merging approach (e.g., task arithmetic) for the non-MLP modules in ViTs while upscaling the MLP modules to a mixture-of-experts (MoE) module equipped with a routing mechanism. The experts in the MoE module consist of task-shared and task-specific knowledge extracted from the fine-tuned MLPs, with the routing mechanism dynamically assigning optimal merge weights to each MLP expert module based on the input samples. 

Although WEMoE exhibits excellent MTL performance, it has potential limitations when upgrading the MLPs in each Transformer block to MoE modules: (1) Dynamically merging MLPs requires additional storage for the MLP parameters across all tasks, leading to a substantial storage burden when the number of tasks is large. (2) Maintaining separate trainable routing parameters for each Transformer block can significantly increase the number of parameters as the number of ViT blocks grows. To address these challenges, we build on WEMoE by introducing an \textbf{E}fficient-and-effective \textbf{W}eight-\textbf{E}nsembling \textbf{M}ixture \textbf{o}f \textbf{E}xperts (E-WEMoE) to perform multi-task model merging. For the first challenge, E-WEMoE eliminates most of the lower-magnitude elements in the task vectors corresponding to the task-specific MLPs in the MoE module. 
For the second challenge, we reformalize the routing task of each Transfermer block as a form of MTL, that is, dynamically allocate the merger weight of the MoE in each block based on a routing strategy shared across blocks.

We conduct extensive experiments to verify the effectiveness of our methods. Specifically, we merged eight benchmark tasks across three ViT architectures. Experimental results demonstrate that our WEMoE and E-WEMoE methods achieve SOTA performance in multi-task model merging settings, reaching levels comparable to traditional multi-task joint training (in Figure~\ref{fig:scatter}). Additionally, we demonstrate the exceptional generalization capabilities of WEMoE and E-WEMoE on unseen tasks, as well as their robustness in scenarios involving distribution drift.

To summarize, \textbf{our contributions} are as follows:
\begin{itemize}
[noitemsep,topsep=0pt,parsep=0pt,partopsep=0pt,leftmargin=*]
    \item We reveal the differences in parameter changes before and after fine-tuning the core modules of Transformer-based ViT models, establishing critical and non-critical modules related to downstream tasks. 
    \item We propose a novel WEMoE model to construct a unified multi-task model, that statically merges non-critical modules and dynamically merges MoE modules upscaled by critical modules based on input samples.
    \item We propose an efficient and effective WEMoE model (E-WEMoE) that reduces both the total parameter cost and the number of trainable parameters by pruning non-essential elements in critical modules and sharing routing across the Transformer blocks.
    \item We conduct extensive experiments to demonstrate the excellent MTL performance, generalization ability and robustness of our WEMoE and E-WEMoE methods.
\end{itemize}

\section{Revisiting Model Merge for MTL}
\label{sec:revisting}

This section first introduces the problem setting and notations in Section~\ref{subsection:problem_formulation}. Then, we revisit the model merge methods for MTL and discuss their limitations in Section~\ref{subsection:revisiting_model_merge}. Finally, we discuss how to extract and split the task-shared and task-private knowledge from the fine-tuned model in Section~\ref{subsection:Knowledge_separation}.

\subsection{Problem Formulation}
\label{subsection:problem_formulation}

Consider a large-scale Transformer-based~\cite{vaswani2017attention} pre-trained model $f_{\theta_0}$, parameterized by $\theta_0 \in \mathbb{R}^{|\theta|}$, where $|\theta|$ denotes the number of parameters. Without loss of generality, this paper focuses on the Visual Transformer (ViT)~\cite{dosovitskiy2020vit} architectures for handling visual tasks.
In the model merging based MTL scenario, a series of downstream tasks $\mathcal{S} = \{s_i\}_{i=1}^n$ are used to fine-tune the pre-trained model $f_{\theta_0}$ independently. The fine-tuned models are denoted as $f_{\theta_i}$, where $\theta_i$ represents the fine-tuned parameters of the model for task $s_i$.

Based on the above notations, we present the definitions of model merging and task vector in Definitions \ref{def:merging} and \ref{def:taskverctor}, respectively. Building on these, we introduce the concept of task vector-based model merging in Definition \ref{def:merging_tv}, which is essential for the subsequent content of this paper.

\begin{definition}[Model Merging]
\label{def:merging}
Model merging aims to \texttt{merge} multiple fine-tuned models $\{f_{\theta_i}\}_{i=1}^n$ at the \textit{parameter level} to create a single model $f_{\theta_{\text{merged}}}$ that can effectively handle all downstream tasks $\{s_i\}_{i=1}^n$.
    \begin{equation}
    \setlength{\abovedisplayskip}{3pt} 
    \setlength{\abovedisplayskip}{3pt} 
        \label{eq:merging}
        f_{\theta_{\text{merged}}} = \texttt{merge}(f_{\theta_1},f_{\theta_2}, \ldots, f_{\theta_n})
      \end{equation}
\end{definition}

\begin{definition}[Task Vector~\citep{ilharcoEditingModelsTask2023}]
\label{def:taskverctor}
    The task vector $\tau_i$ is defined as the difference between the parameters of the fine-tuned model $f_{\theta_i}$ and those of the pre-trained model $f_{\theta_0}$, i.e.,
      \begin{equation} \setlength{\abovedisplayskip}{3pt} 
    \setlength{\abovedisplayskip}{3pt} 
        \label{eq:task_vector}
        \tau_i = \theta_i - \theta_0.
      \end{equation}
\end{definition}

\begin{definition}[Task Vector based Model Merging]
\label{def:merging_tv}
Task vector-based model merging involves to \texttt{merge} multiple task-specific vectors $\{\tau_i\}_{i=1}^n$ into the pre-trained model $f_{\theta_0}$, enabling the merged model $f_{\theta_{\text{merged}}}$ to address all downstream tasks.
       \begin{equation} \setlength{\abovedisplayskip}{3pt} 
    \setlength{\abovedisplayskip}{3pt} 
        \label{eq:merging_tv}
        f_{\theta_{\text{merged}}} = \texttt{merge}({\theta_0}, \tau_1, \ldots, \tau_n)
      \end{equation}
\end{definition}

While there is a wealth of advanced work on model merging based on task vectors~\cite{Survery_ModelMerging_2024,yadavResolvingInterferenceWhen2023,yangAdaMergingAdaptiveModel2023,yu2024language,wang2024localizing}, most primarily focus on \textit{static} merging models. In the next subsection, we will discuss the limitations of this static merging approach.

\subsection{Revisiting Static Model Merging}
\label{subsection:revisiting_model_merge}

From the perspective of multi-objective optimization, the solution space for the optimal merged model $f_{\theta_{\text{merged}}}$ should be represented by the Pareto front constructed from all downstream tasks in $\mathcal{S}$. In other words, an optimal merged model is expected to simultaneously achieve the best performance across all tasks in $\mathcal{S}$.
Unfortunately, identifying an optimal model in the model merging context is quite challenging for several reasons: (i) the solution space of the merged model is large and complex; and (ii) the model merging approach only has access to the trained and fine-tuned models, without access to the training data of the downstream task. This limitation makes it impossible to train the optimal model from scratch using a multi-objective optimization algorithm, as is done in traditional methods.

\begin{figure}[t]
  \begin{center}
    \includegraphics[width=.95\linewidth]{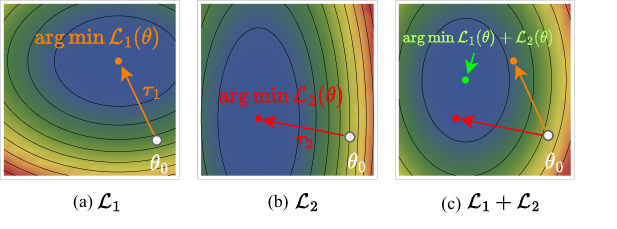}
    \vskip -0.1in
    \caption{
      An illustration of the loss landscapes of $s_1$, $s_2$, and $s_1 \cup s_2$.
      There is no static solution $\theta'$ that simultaneously minimizes the loss of both tasks better than $\argmin_\theta \mathcal{L}_1(\theta) + \mathcal{L}_2(\theta)$.
    }
    \label{fig:loss_landscapes}
  \end{center}
  \vskip -0.2in
\end{figure}

Furthermore, a single Pareto optimal solution derived from model merging may still suffer from performance loss compared to the individual expert models. Specifically, the core idea behind Pareto optimal solutions~\cite{sener2018multi} is that a solution is considered Pareto optimal if no alternative solution can simultaneously improve one objective without degrading at least one other objective. For instance, we cannot minimize the loss of all tasks better than the global optimal joint loss function $\small \argmin_\theta \sum_{i=1}^n \mathcal{L}_i(\theta)$, where $\mathcal{L}_i$ represents the loss function of a single downstream task $s_i$.
An example of the loss landscape for tasks $s_1$, $s_2$, and $s_1 \cup s_2$ is illustrated in Figure~\ref{fig:loss_landscapes}. It is evident that there is no static solution $\theta'$ that satisfies both $\small \mathcal{L}_1(\theta') < \mathcal{L}_1(\theta^{*})$ and $\small \mathcal{L}_2(\theta') < \mathcal{L}_2(\theta^{*})$, where $\small \theta^{*} = \argmin_\theta \mathcal{L}_1(\theta) + \mathcal{L}_2(\theta)$. However, $\theta^*$ is a suboptimal solution for the two tasks individually, as shown by $\small \mathcal{L}_1(\theta^*) > \min_\theta \mathcal{L}_1(\theta)$ and $\small \mathcal{L}_2(\theta^*) > \min_\theta \mathcal{L}_2(\theta)$. In Figure~\ref{fig:loss_landscapes_examples} of Appendix~\ref{appendix:loss}, we provide loss landscapes on multiple real-world datasets.

In summary, obtaining an ideal Pareto solution through model merging presents significant challenges. Furthermore, a single Pareto solution is not necessarily optimal compared to an individual expert model. Current task vector-based merging methods combine task vectors into the pre-trained model using various strategies to produce a single \textit{static} merged model. However, this static approach restricts the adaptability of the merged model to the specific requirements of individual input instances, thereby limiting its performance. We argue that an ``ideal'' merged model should \textit{dynamically} merge shared knowledge between tasks with task-specific knowledge, tailored to each input instance.

\begin{figure*}[t]
 \vskip -0.1in
  \begin{center}
    \centerline{\includegraphics[width=.9\linewidth]{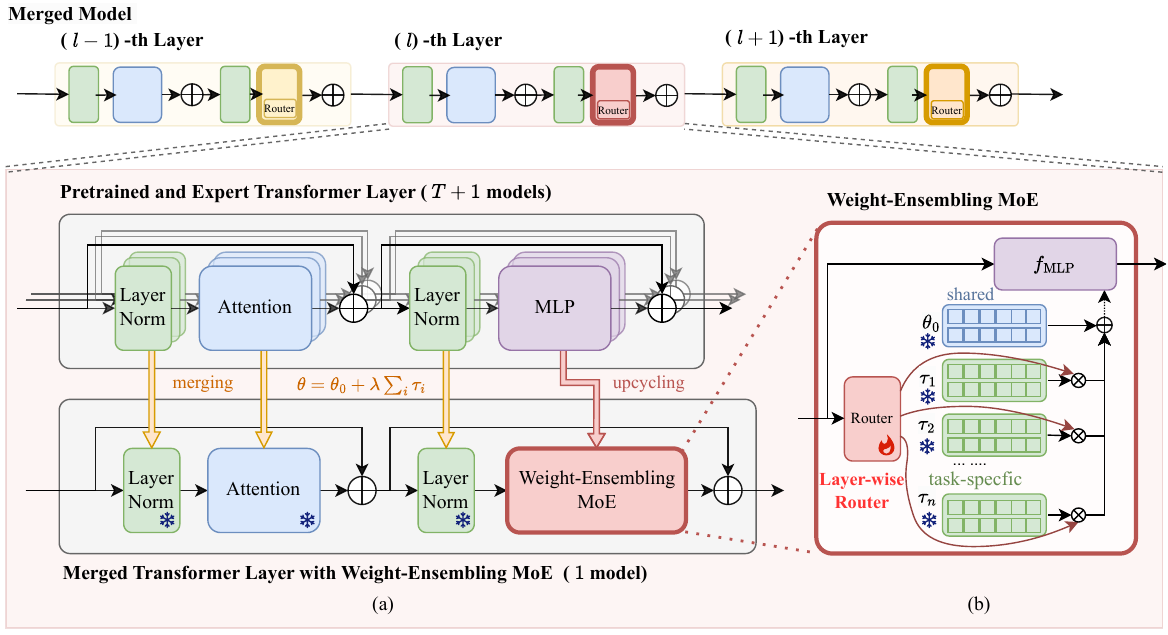}}
    \vskip -0.05in
    \caption{
      (a) \textbf{Overview of the Weight-Ensembling Mixture of Experts (WEMoE) Framework}. This figure illustrates the overall framework of our proposed approach for merging the pre-trained model with fine-tuned task-specific models. We perform weight merging across the Transformer layers, excluding the MLPs.  For the MLPs, we upcycle them into weight-assembling MoE modules.
      (b) \textbf{WEMoE module}. This diagram details the structure of the WEMoE module, which consists of a router, the pre-trained MLP weights, and a set of task vectors w.r.t. MLP modules.
    }
    \label{fig:overview}
  \end{center}
  \vskip -0.2in
\end{figure*}

\subsection{Knowledge Extraction and Separation}
\label{subsection:Knowledge_separation}

As mentioned in the previous section, to construct an ideal merging model, it is essential to distinguish between task-private and task-shared knowledge, dynamically merging them based on the specific needs of each instance. 

Knowledge splitting is a crucial step in constructing an ideal merged model. Common methods for knowledge separation include knowledge distillation~\citep{hintonDistillingKnowledgeNeural2015}, pruning~\citep{franklePruningNeuralNetworks2021}, and feature extraction~\citep{yosinskiUnderstandingNeuralNetworks2015}. However, these traditional strategies often involve significant computational overhead or data dependence when separating or extracting knowledge from deep neural networks, rendering them unsuitable for model merging scenarios. Therefore, our objective is to separate shared knowledge and task-specific knowledge in a computationally efficient and data-free manner. Fortunately, according to the definition of the task vector (in Definition \ref{def:taskverctor}), it can be considered a carrier of task-specific knowledge, as $\tau_i$ measures the difference in parameter changes after fine-tuning on the downstream task $s_i$. Meanwhile, the knowledge contained in the pre-trained model $\theta_0$ can be viewed as knowledge shared across tasks.

However, there are significant limitations to simply treating whole $\theta_0$ as shared knowledge and whole $\tau_i$ as task-private knowledge, then dynamically merging models based on inputs during inference. This approach necessitates storing the pre-trained model and all individual expert models, which contradicts the primary goal of efficient MTL: to achieve a parameter scale for the MTL model that is comparable to that of a single model, without maintaining a separate set of parameters for each task. Therefore, we argue that a more fine-grained separation between shared knowledge and task-private knowledge is essential, with shared knowledge occupying most of the parameters while minimizing the number of parameters allocated to task-private knowledge.

\begin{figure}[h]
    \centering
    \begin{minipage}{0.24\textwidth}
        \centering
        \includegraphics[width=\linewidth]{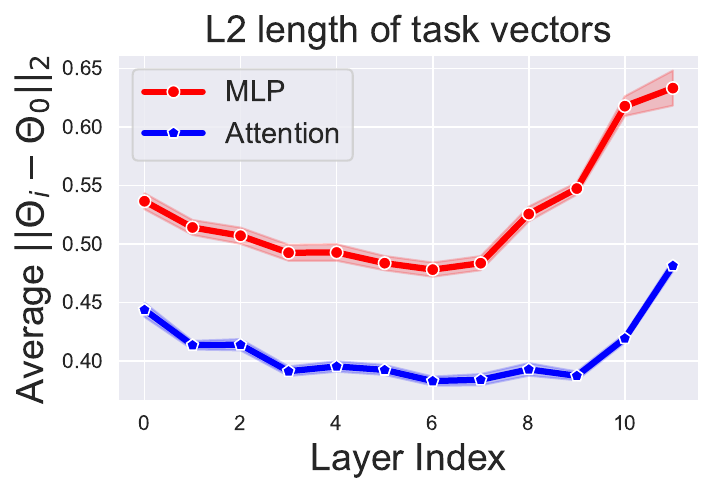}
    \end{minipage}
    \hfill
      \begin{minipage}{0.24\textwidth}
        \centering
        \includegraphics[width=\linewidth]{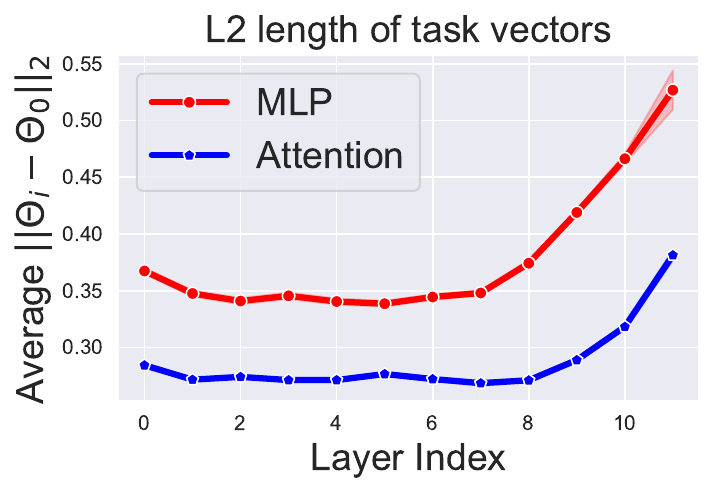}
    \end{minipage}
    \hfill
    \vskip -0.1in
    \caption{The distance between the parameters of the pre-trained model and the fine-tuned models. 
    The first sub-figure shows the average $L_2$ distance of CLIP-ViT-B/32 on eight datasets, and the last sub-figure is on CLIP-ViT-B/16.
    }
    \label{fig:distance_between_parameters}
\vskip -0.2in
\end{figure}

In this paper, to achieve a fine-grained split between shared and private knowledge, we revisit the key components in the Transfermer block of ViT, specifically the Multi-Head Self-Attention (Att) and Feed-Forward Neural Network (MLP). 
Let $\small \theta_{l,i}^{mlp}$ and $\small \theta_{l,i}^{att}$ denote the parameters of the MLP and Att modules, respectively, in the $l$-th ($\small l \in \{1,2,\ldots,L\}$) block of the fine-tuned model $\small f_{\theta_i}$ w.r.t. task $s_i$, where $L$ represents the total number of blocks. Similarly, $\small \theta_{l,0}^{mlp}$ and $\small \theta_{l,0}^{att}$ denote the parameters of the MLP and Att modules in the pre-trained model $\small f_{\theta_0}$. We then measure the difference between $\small \theta_{l,i}^{mlp}$ and $\small \theta_{l,0}^{mlp}$ (or $\theta_{l,i}^{att}$ and $\small \theta_{l,0}^{att}$) using the $L_2$ distance, which quantifies the magnitude of parameter changes before and after fine-tuning. The formal statement is $L_2^{mlp} = \small || \theta_{l,i}^{mlp}-\theta_{l,0}^{mlp} ||_2^2$, $L_2^{att} = \small || \theta_{l,i}^{att}-\theta_{l,0}^{att} ||_2^2$.
Intuitively, a smaller $L_2^{mlp}$ or $L_2^{att}$ distance indicates that the downstream task $s_i$ causes minimal alterations to the pre-trained knowledge (i.e., $\theta_{l,0}^{att}$ or $\theta_{l,0}^{mlp}$), suggesting that these parameters are more easily shared across tasks without conflict. In contrast, a larger distance $L_2^{mlp}$ or $L_2^{att}$ signifies a stronger association with the downstream task, increasing the likelihood of task conflicts during model merging. 
As shown in Figure~\ref{fig:distance_between_parameters}, we analyzed eight benchmark datasets using the ViT-B/32 (left) and ViT-B/16 (right) architectures, reporting the $L_2^{mlp}$ and $L_2^{att}$ distances for the Att and MLP modules in each Transfermer block (or layer). The results consistently show that the MLP module exhibits a greater magnitude of change compared to the Att module, indicating a higher likelihood of conflicts in the MLP modules. Therefore, we designate the MLP modules as crucial modules sensitive to the downstream task, while other non-MLP modules are considered as non-crucial modules.

In the following section, we will discuss how to dynamically merge the crucial modules based on input instances while statically merging the non-key modules to strike an optimal balance between MTL performance and memory efficiency.

\section{Methodology}
\label{section:methodology}

In this section, we first present the overall framework of the proposed approach in Section \ref{subsection:framework_overview}, followed by an explanation of how the MLP module is upcycling to a WEMoE structure in Section \ref{subsection:weight_ensembling_moe_module}. In Section \ref{subsec:ewemoe}, we introduce the efficient and effective E-WEMoE model built upon WEMoE.

\subsection{Framework Overview}
\label{subsection:framework_overview}

The overview framework of our proposed approach is shown in Figure~\ref{fig:overview}(a). The core insight of our approach is to separate task-shared and task-specific knowledge from multiple fine-tuned expert models and dynamically merge the crucial modules most prone to conflict based on the input. In other words, the main advantage of our framework lies in its ability to dynamically integrate shared knowledge and task-specific information in response to input samples, rather than relying on a static solution within the original parameter space. This dynamic adjustment enables the merged model to better adapt to the fine-grained differences of each task or individual sample, thereby enhancing its performance across the source tasks.

Based on our analysis in Section \ref{subsection:Knowledge_separation}, we consider all non-critical modules (i.e, non-MLP modules in each Transformer block) as knowledge that can be easily shared across tasks and perform static merging to reduce the total number of parameters in the merged model. Without loss of generality, we employ the classic Task Arithmetic~\cite{ilharcoEditingModelsTask2023} to merge non-critical modules, adding the sum of the task vectors to the pre-trained model through a predefined merging coefficient, $\lambda$, namely:
\begin{equation}
\small
 \setlength{\abovedisplayskip}{3pt} 
    \setlength{\abovedisplayskip}{3pt} 
    \theta_{merged, l}^{h} = \theta_{0,l}^{h} + \lambda \sum_{i=1}^n \tau_{i,l}^{h}, \;\; \forall \; l \in \{1,2,\ldots,L\}
\label{eq:taskarithmetic}
\end{equation}
where $ h \in$ \{Attention, Layer Norm\} represents all non-critical modules, and $l$ represents the $l$-th Tranformer block.

For the critical modules (i.e, MLP modules in each Transformer block), we upgrade them to a WEMoE structure, which will be introduced in Section \ref{subsection:weight_ensembling_moe_module}. The merging goal of WEMoE is formally stated as:
\begin{equation}
\small
 \setlength{\abovedisplayskip}{3pt} 
    \setlength{\abovedisplayskip}{3pt} 
  \begin{split}
        \theta_{merged, l}^{mlp} &= \texttt{WEMoE}(\theta_{0,l}^{mlp}, \tau_{1,l}^{mlp},\tau_{2,l}^{mlp},\ldots,\tau_{n,l}^{mlp} | \texttt{R}_l) \;\; 
  \end{split}
\label{eq:moemlp}
\end{equation}
where $\forall \; l \in \{1,2,\ldots,L\}$, and $ \texttt{R}_l$ is the router.

\subsection{Weight-Ensembling MoE Module}
\label{subsection:weight_ensembling_moe_module}

In this subsection, we provide a detailed explanation of the Weighted Ensemble Mixture of Experts (WEMoE) module. As illustrated in Figure \ref{fig:overview} (b), the WEMoE module in $l$-th Transformer block consists of three main components: a router $\texttt{R}_l$, the pre-trained MLP weight $\theta_{0,l}^{mlp}$, and a set of task vectors $\{\tau_{1,l}^{mlp}, \tau_{2,l}^{mlp},\ldots,\tau_{n,l}^{mlp}\}$ associated with the MLP for the downstream tasks. 

The role of the router, $\texttt{R}_l:$ $\mathbb{R}^d \rightarrow \mathbb{R}^n$, is to generate the merged weights $\mathbf{\lambda}_l=\{\lambda_{1,l},\lambda_{2,l},\ldots,\lambda_{n,l}\}$ in the $l$-th block by combining the pre-trained weights and the expert task vector based on the input features $\mathbf{h}^{in}_l$ of the instance at the $l$-th block. The router's implementation is flexible and can be adapted to various architectures. In this paper, we employ a straightforward multi-layer fully connected network to implement it, without loss of generality. The mathematical formulation of the WEMoE module is given by:
\begin{align}
\small
 \setlength{\abovedisplayskip}{3pt} 
    \setlength{\abovedisplayskip}{3pt} 
  \label{eq:routing_weights}
  \texttt{WEMoE} \left\{
  \begin{array}{l}
    \mathbf{\lambda}_l = \texttt{mean}(\texttt{R}_l(\mathbf{h}_{1:N}^{{in,l}})) \in \mathbb{R}^{n \times 1}, \\
    \theta^{{mlp}}_{merged, l} = \theta_{0,l}^{{mlp}} + \mathbf{D}^{\tau}_l \cdot \mathbf{\lambda}_l,             \\
    \mathbf{h}_{1:N}^{{out,l}} = f_{{MLP}}(\mathbf{h}_{1:N}^{{in,l}}; \theta^{{mlp}}_{merged, l}).
  \end{array}
  \right.
\end{align}
where $\small \mathbf{h}_{1:N}^{in,l}$ represents the input sequence of tokens at $l$-th block, $\small \mathbf{h}_{1:N}^{{out},l}$ denotes the output sequence of tokens, $f_{MLP}(\cdot)$ is the MLP function, and $\small \mathbf{D}_{\tau}^l \in \mathbb{R}^{|\theta_{0,l}^{{mlp}}| \times n}$ is the dictionary matrix of task vectors, composed of fixed task vectors, that is, $\small \mathbf{D}_{\tau}^l = [\tau_{1,l}^{mlp}; \tau_{2,l}^{mlp};\ldots;\tau_{n,l}^{mlp}]$. The $\texttt{mean}(\cdot)$ function averages the routing weights across all tokens in the input sequence.
From a dictionary learning perspective, the WEMoE module can be interpreted as a dictionary lookup operation, where the routing weights are used to select task vectors from the dictionary matrix $\mathbf{D}^{\tau}_l$. These selected vectors are then added to the pre-trained MLP weights $\theta_{0,l}^{mlp}$, resulting in input-conditioned MLP weights $\theta_l^{{mlp}}$.
Next, we will detail the design, initialization, and fine-tuning of the router's parameters.

\subsubsection{The Structure of Router}
Routing is a core component of WEMoE, and its design and initialization play a crucial role in the merged model's performance. In this paper, we implement routing using multiple fully connected layers, focusing on two main configurations: one with no hidden layers ($l_{fc}=0$) and another with two hidden layers ($l_{fc}=2$). Additional configurations with different numbers of hidden layers are analyzed in Appendix \ref{appendix:ablations_of_router_depth}. The mathematical formulation for routing under these configurations is as follows:
\begin{equation}
\small
 \setlength{\abovedisplayskip}{3pt} 
    \setlength{\abovedisplayskip}{3pt} 
  \texttt{R}_l(h) = 
  \left\{
  \begin{array}{l}
    \mathbf{W}_1 \texttt{ReLU}(\mathbf{W}_0 h + \mathbf{b}_0) + \mathbf{b}_1, \; \text{if} \; l_{fc}=2, \\
    \mathbf{b}_0, \; \text{if} \; l_{fc}=0.
  \end{array}
  \right.
\label{eq:routing_depth_wemoe}
\end{equation}
where $\small \mathbf{W}=\{\mathbf{W}_1, \mathbf{W}_0\}$ and $\small \mathbf{b}=\{\mathbf{b}_1, \mathbf{b}_0\}$ represent the weight and bias parameters, respectively, and $\texttt{ReLU}(\cdot)$ is the rectified linear unit activation function. Unless stated otherwise, we adopt the structure with $l_{fc}=2$.

\subsubsection{The Initialization of Router}
The initialization of the router significantly impacts the performance of the merged model. To ensure that the initial routing weight is close to $\lambda$ (the merging coefficient in Task Arithmetic in Eq.~\ref{eq:taskarithmetic}), we initialize $\mathbf{W}_0$ and $\mathbf{W}_1$ by sampling from a Gaussian distribution with a mean of 0 and a variance of 0.01, and set $\mathbf{b}_0$ to 0 and $\mathbf{b}_1$ to $\lambda$. When $l=0$, we directly assign $\lambda$ to $\mathbf{b}_0$. For the case where $l_{fc}=0$, the router can be viewed as a partial implementation of AdaMerging \citep{yangAdaMergingAdaptiveModel2023}, in which only the critical MLP module performs adaptive merging coefficient optimization, while the other modules use the preset $\lambda$ from Task Arithmetic.

\subsubsection{Training Router via Test-Time Adaptation}
\label{subsection:test_time_adaptation_training}

As shown in Eq.~\ref{eq:routing_depth_wemoe}, the router includes the parameters $\mathbf{W}$ and $\mathbf{b}$ that require tuning. Once the router is initialized, the next step is to fine-tune its parameters. However, in the context of model merging, we do not have access to the original training data for the downstream tasks. To address this issue, we employ test-time adaptation training techniques, which are widely utilized in the field of semi-supervised learning~\citep{mounsavengBagTricksFully2023,liangComprehensiveSurveyTestTime2023}. Test-time adaptation is a powerful approach that enables the model to adjust its parameters based on unlabeled test data during the testing phase, thereby enhancing the performance of the merged model. In this research, we concentrate on classification tasks and aim to minimize the multi-task entropy loss of the merged model using unlabeled test data like AdaMerging~\cite{yangAdaMergingAdaptiveModel2023}. Entropy loss serves as a measure of prediction uncertainty and is defined as follows:
\begin{align}
 \setlength{\abovedisplayskip}{3pt} 
    \setlength{\abovedisplayskip}{3pt} 
\label{eq:entropy_loss}
\small
  \mathcal{L}_{\text{entropy}}
   & = \mathbb{E}_{x\sim \mathcal{D}_{\text{test}}}[-p(\hat{y}|x) \log p(\hat{y}|x)]           \\
   & \approx -\frac{1}{|D|}\sum_{i=1}^{|D|} \sum_{c=1}^C p(\hat{y}_c|x_i) \log p(\hat{y}_c|x).
\end{align}
where $\mathcal{D}_{\text{test}}$ represents the test data distribution, $p(\hat{y}|x)$ denotes the predicted posterior probability distribution, $C$ is the total number of categories, and $|D|$ is the number of samples in the test dataset. This optimization objective encourages the merged model to generate predictions with greater confidence, thereby enhancing the prediction performance on the test data. When the model is confident in its decisions, it is more likely to produce accurate predictions.

\begin{figure*}[t]
  \begin{center}
    \centerline{\includegraphics[width=.9\linewidth]{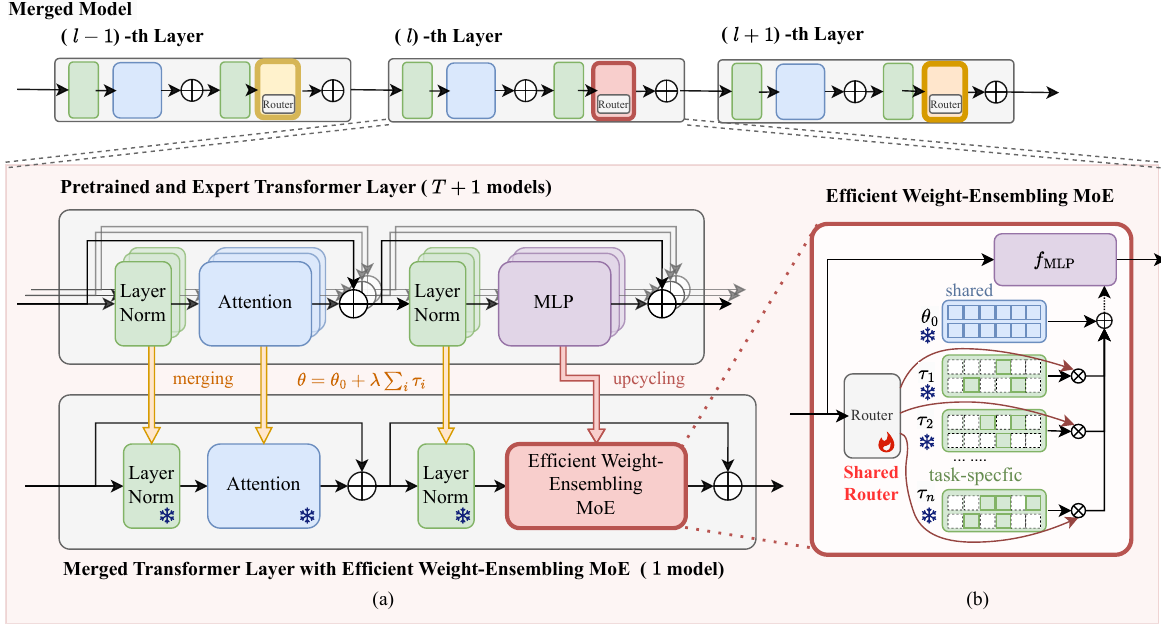}}
    \caption{
      (a) \textbf{Overview of the Efficient Weight-Ensembling Mixture of Experts (E-WEMoE) Framework}. It merges all non-MLP modules through task arithmetic and upgrades the MLP modules into an efficient E-WEMoE module. 
      (b) \textbf{E-WEMoE Module}. The module includes a router shared across all Transformer blocks, the pre-trained MLP module, and a set of sparse task-specific vectors w.r.t. MLP modules.
    }
    \label{fig:overview_v2}
  \end{center}
  \vskip -0.1in
\end{figure*}

\begin{figure*}[t]
  \begin{center}
  \centerline{\includegraphics[width=0.9\linewidth]{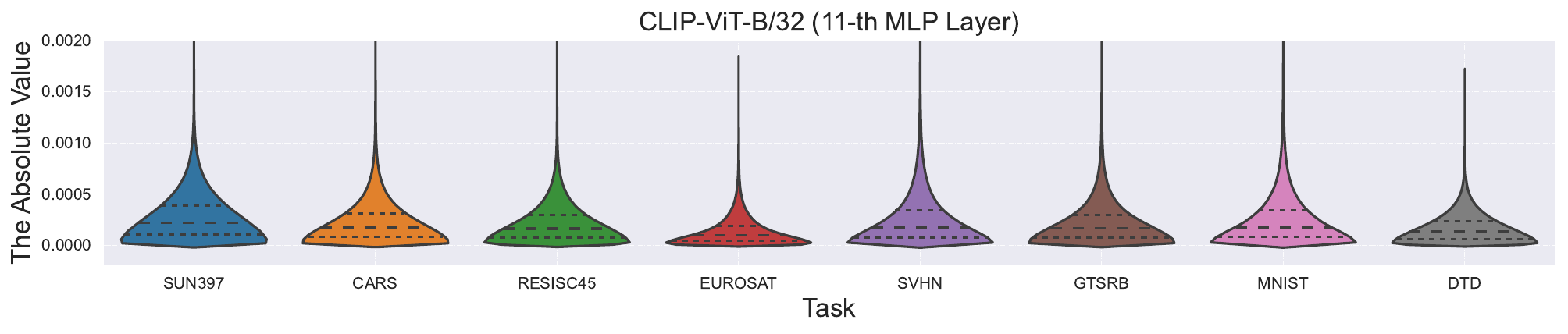}}
 \vskip -0.1in
 \caption{
    An illustration of the element magnitudes in the task vector constructed by the MLP in the 11-th Transformer block of ViT-B/32. Additional layers are presented in Figure~\ref{fig:violinplot_vitb32_appendix} of the appendix.
    }
   \label{fig:violinplot_vitb32_layer11}
  \end{center}
  \vskip -0.3in
\end{figure*}

\textbf{Discussion on the Advantages and Limitations of WEMoE}.
\label{subsec:discussion_wemoe}
In summary, our WEMoE method offers two key \textbf{\textit{advantages}}:  
(1) It incorporates cross-task shared knowledge, represented by the weights $\small \{\theta_{0,l}^{mlp}\}_{l=1}^L$ of the MLP in the pre-trained model, as well as task-private knowledge, encapsulated in the task vector $\small \{\tau_{t,l}^{mlp}\}_{t=1,l=1}^{n,L}$ derived from the MLP weights of the expert models.  
(2) It enables the dynamic integration of private knowledge and diverse expert knowledge based on the input samples, a process regulated by the routing mechanism.
However, compared to traditional static model merging methods, our WEMoE approach has two \textbf{\textit{limitations}}:
(1) It requires the storage of the MLP modules for all downstream tasks. When the number of tasks is large, and the MLP modules contain many elements, WEMoE imposes a significant storage burden on the merged model.
(2) It includes a routing mechanism in each Transformer block, which contains parameters that require fine-tuning. For larger-scale ViT models (i.e., those with more Transformer blocks), the total number of tuning parameters increases substantially.

\subsection{Efficient Weight-Ensembling MoE (E-WEMoE)}
\label{subsec:ewemoe}

It is crucial to retain the two advantages of WEMoE while mitigating its two limitations; otherwise, its applicability will be restricted. In this subsection, we discuss how to implement an efficient and effective version of WEMoE. Specifically, we begin with an in-depth analysis of the two limitations discussed in Section \ref{subsection:weight_ensembling_moe_module}, then present the overall E-WEMoE framework and a detailed explanation of its components.

\subsubsection{Revisiting the Task Vectors w.r.t. MLP in WEMoE}
\label{subsubsec:revisiting_mlp_tv}
Each MLP module within a Transformer block comprises two linear layers and a nonlinear activation function. For instance, in CLIP-ViT-B/32, the task vectors w.r.t. MLP module's parameters consist of $\small \tau_{t,l}^{mlp} = \{\mathbf{W}_1 \in \mathbb{R}^{768 \times 3072}, \mathbf{b}_1 \in \mathbb{R}^{3072}, \mathbf{W}_0 \in \mathbb{R}^{3072\times 768}, \mathbf{b}_0 \in \mathbb{R}^{768}\}$. When merging $n$ ViT models, each with $L$ blocks, the additional parameter amount is $N_p = n\times L\times |\tau_{t,l}^{mlp}|$.
Interestingly, a statistical analysis of the magnitudes of the task vectors for each MLP reveals that most of the values in $\tau_{t,l}^{mlp}$ are very small, approaching zero. 
Fig.~\ref{fig:violinplot_vitb32_layer11} illustrates this using a violin plot, which shows the distribution of task vector magnitudes for eight tasks in the $11$-th block. We observed that the absolute values of over 75\% (the highest dotted line) of the elements are less than 0.0005, with the median values (the middle dotted line) of all elements ranging between 0.0002 and 0.0003. Similar patterns are also evident across additional layers in the Fig.~\ref{fig:violinplot_vitb32_appendix} of the appendix. This observation inspires us to mask the majority of elements in task vectors $\small \{\tau_{t,l}^{mlp}\}_{t=1,l=1}^{n,L}$, thus conserving storage space for the task vectors.

\subsubsection{Revisiting the Router in WEMoE}
\label{subsubsec:revisiting_mlp_router}
Recall that the role of the router $\mathbb{R}_l$ in the WEMoE module in Section \ref{subsection:weight_ensembling_moe_module} is to receive the layer's input hidden feature $h^{in,l}$ and output the merging weights $\mathbf{\lambda}_l$ of the layer, formally expressed as:
\begin{equation}
\small
 \setlength{\abovedisplayskip}{3pt} 
    \setlength{\abovedisplayskip}{3pt} 
\text{
\parbox{1.cm}{\centering \texttt{Layer}\\\texttt{Wise}\\\texttt{Router}}
}
\left\{
  \begin{array}{l}
\mathbf{\lambda}_1 = \texttt{mean}(\texttt{R}_1(\mathbf{h}_{1:N}^{{in,1}})) \in \mathbb{R}^{n \times 1}, \text{ if } l\!=\!1 \\
\mathbf{\lambda}_2 = \texttt{mean}(\texttt{R}_2(\mathbf{h}_{1:N}^{{in,2}})) \in \mathbb{R}^{n \times 1}, \text{ if } l\!=\!2 \\
   \hfill \vdots \hfill \\
\mathbf{\lambda}_L = \texttt{mean}(\texttt{R}_L(\mathbf{h}_{1:N}^{{in,L}})) \in \mathbb{R}^{n \times 1}, \text{ if } l\!=\!L \\
  \end{array}
\right.
\end{equation}
As shown in Eq.~\ref{eq:routing_weights}, routers $\texttt{R}_1$, $\texttt{R}_2$, and $\texttt{R}_L$ contain parameters that need to be fine-tuned. When the number of layers $L$ is large, the amount of parameters introduced cannot be ignored.
Interestingly, in the Transformer-based architecture, we found that the input features of each layer maintain a consistent shape, specifically $\small \texttt{shape}(\mathbf{h}_{1:N}^{{in,1}}) = \texttt{shape}(\mathbf{h}_{1:N}^{{in,2}}) =\cdots = \texttt{shape}(\mathbf{h}_{1:N}^{{in,L}})$. Furthermore, the output of each router also exhibits a consistent shape, such that $\small \texttt{shape}(\mathbf{\lambda}_1) = \texttt{shape}(\mathbf{\lambda}_2)=\cdots= \texttt{shape}(\mathbf{\lambda}_L)$. This motivates exploring whether the idea of MTL can be used to share a cross-layer router $\texttt{R}_{shared}$ to reduce the number of trainable parameters.

\textbf{Framework Overview}.
Based on the aforementioned motivations, we propose an efficient and effective WEMoE model, referred to as E-WEMoE. This model sparsifies the task vectors of the MLP and employs cross-layer shared routing to reduce both the total number of parameters in the merged model and the trainable parameters of the router. The overall framework of E-WEMoE is illustrated in Fig.~\ref{fig:overview_v2}(a). Similar to WEMoE, E-WEMoE utilizes static task arithmetic  (i.e., Eq.~\ref{eq:taskarithmetic}) merging for the parameters of non-MLP modules, while upgrading the MLP parameters to an E-WEMoE module, that is, 
\begin{equation}
\small
 \setlength{\abovedisplayskip}{3pt} 
    \setlength{\abovedisplayskip}{3pt} 
  \begin{split}
        \theta_{merged, l}^{mlp} &= \texttt{E-WEMoE}(\theta_{0,l}^{mlp}, \tau_{1,l}^{mlp},\tau_{2,l}^{mlp},\ldots,\tau_{n,l}^{mlp}  | \texttt{R}_{shared}) \;\; 
  \end{split}
\label{eq:moemlp}
\end{equation}
where $\forall \; l \in \{1,2,\ldots,L\}$, and $\texttt{R}_{shared}$ is the shared router. 

\textbf{E-WEMoE module}. In this part, we explain the implementation details of the E-WEMoE module, shown in Fig.~\ref{fig:overview_v2}(b). First, we prune the task vector $\tau_{t,l}^{mlp}$ based on the magnitude of its elements, using a hyperparameter $\rho$ (we set $\rho$ to $90\%$ by default) to control the sparsity ratio. For instance, E-WEMoE-$90\%$ indicates that $90\%$ of the elements in $\tau_{t,l}^{mlp}$ are removed. Therefore, the dictionary structure of E-WEMoE is: 
\begin{equation}
\small
 \setlength{\abovedisplayskip}{3pt} 
    \setlength{\abovedisplayskip}{3pt} 
    \texttt{sparse}(\mathbf{D}^{\tau}_l, \rho) = [\texttt{sparse}(\tau_{1,l}^{mlp}, \rho),\ldots, \texttt{sparse}(\tau_{L,l}^{mlp}, \rho)]
\end{equation}
Next, our routing mechanism in E-WEMoE is represented as:
\begin{equation}
\small
 \setlength{\abovedisplayskip}{3pt} 
    \setlength{\abovedisplayskip}{3pt} 
\text{
\parbox{1.cm}{\centering \texttt{Shared}\\\texttt{Router}}
}
\left\{
  \begin{array}{l}
\mathbf{\lambda}_1 = \texttt{mean}(\texttt{R}_{shared}(\mathbf{h}_{1:N}^{{in,1}})) \in \mathbb{R}^{n \times 1}, \text{ if } l\!=\!1 \\
\mathbf{\lambda}_2 = \texttt{mean}(\texttt{R}_{shared}(\mathbf{h}_{1:N}^{{in,2}})) \in \mathbb{R}^{n \times 1}, \text{ if } l\!=\!2 \\
   \hfill \vdots \hfill \\
\mathbf{\lambda}_{L} = \texttt{mean}(\texttt{R}_{shared}(\mathbf{h}_{1:N}^{{in,L}})) \in \mathbb{R}^{n \times 1}, \text{ if } l\!=\!L \\
  \end{array}
\right.
\end{equation}
To summarize, the complete computation rule for E-WEMoE is given by:
\begin{align}
\small
 \setlength{\abovedisplayskip}{3pt} 
    \setlength{\abovedisplayskip}{3pt} 
  \label{eq:routing_weights_ewemoe}
  \texttt{E-WEMoE}\!-\!\rho
  \left\{
  \begin{array}{l}
    \mathbf{\lambda}_l = \texttt{mean}(\texttt{R}_{shared}(\mathbf{h}_{1:N}^{{in,l}})) \in \mathbb{R}^{n \times 1}, \\
    \theta^{{mlp}}_{merged, l} = \theta_{0,l}^{{mlp}} + \texttt{sparse}(\mathbf{D}^{\tau}_l, \rho) \cdot \mathbf{\lambda}_l,             \\
    \mathbf{h}_{1:N}^{{out,l}} = f_{{MLP}}(\mathbf{h}_{1:N}^{{in,l}}; \theta^{{mlp}}_{merged, l}),
  \end{array}
  \right.
\end{align}
where parameters in the shared router $\texttt{R}_{shared}$ are fine-tuned using the test-time adaptation method shown in Section~\ref{subsection:test_time_adaptation_training}.

\begin{figure}[h]
    \centering
     \begin{minipage}{0.24\textwidth}
        \centering
         \includegraphics[width=\linewidth]{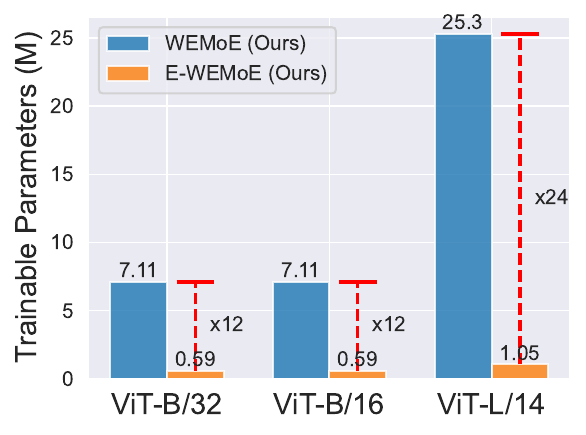}
    \end{minipage}
     \hfill
    \begin{minipage}{0.24\textwidth}
        \centering
        \includegraphics[width=\linewidth]{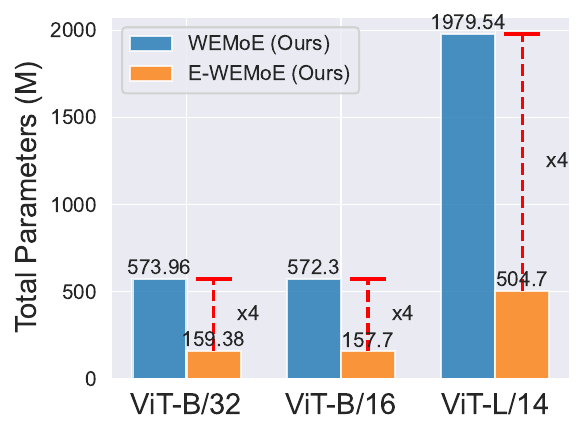}
    \end{minipage}
    \caption{
    Comparison of (a) trainable parameters and (b) total parameters between WEMoE and E-WEMoE-90\%.
    }
    \label{fig:num_pram}
\vskip -0.1in
\end{figure}

\textbf{Discussion on Efficiency and Effectiveness}. We now discuss the efficiency and effectiveness of E-WEMoE in terms of parameter reduction and performance. On the one hand, as illustrated in Fig.~ \ref{fig:num_pram}(a), our shared routing design achieves a significant reduction in the trainable routing parameters for ViT-B/32, ViT-B/16, and ViT-L/14, decreasing them by 12, 12, and 24 times, respectively. Moreover, as shown in Fig.~ \ref{fig:num_pram}(b), the sparsity of the MLP task vector in E-WEMoE-90\% results in a $4\times$ reduction in the total number of parameters for ViT-B/32, ViT-B/16, and ViT-L/14. This shows that our E-WEMoE is more efficient than WEMoE.
On the other hand, as shown in Fig.~\ref{fig:scatter}, E-WEMoE achieves model merging performance comparable to vanilla WEMoE across various sparsity levels $\rho \in \{70\%, 90\%, 95\%, 99\%\}$ for all three architectures, indicating that E-WEMoE remains effective while offering significant parameter reductions. Notably, at $\rho=99\%$, E-WEMoE's parameter scale is nearly identical to that of popular merging methods like Task Arithmetic~\cite{ilharcoEditingModelsTask2023} and AdaMerging~\cite{yangAdaMergingAdaptiveModel2023}, yet E-WEMoE still delivers superior MTL merging performance.

\section{Experiments}
\label{section:experiments}

\begin{table*}[t]
  \caption{Multi-task performance when merging CLIP-ViT-B/32 models on all eight tasks.}
  \label{table:multi-task_performance_clip-vit-b-32}
   \vspace{-10pt}
  \begin{center}
    \setlength{\tabcolsep}{1.5mm}
    \begin{small}
      \begin{sc}
        \begin{tabular}{lccccccccc}
          \toprule
          \textbf{Method}       & \textbf{SUN397} & \textbf{Cars} & \textbf{RESISC45} & \textbf{EuroSAT} & \textbf{SVHN} & \textbf{GTSRB}         & \textbf{MNIST} & \textbf{DTD}    & \textbf{Avg.} \\
          \midrule
          Pre-trained           & 63.2            & 59.6          & 60.2              & 45.0             & 31.6          & 32.6                   & 48.3           & 44.4            & 48.1          \\
          Individual            & \textbf{75.3}   & \textbf{77.7} & \textbf{96.1}     & \textbf{99.9}    & \textbf{97.5} & 98.7                   & \textbf{99.7}  & \textbf{79.4}   & \textbf{90.5} \\
          Traditional MTL       & \copyed{73.9}   & \copyed{74.4} & \copyed{93.9}     & \copyed{98.2}    & \copyed{95.8} & \textbf{\copyed{98.9}} & \copyed{99.5}  & \copyed{77.9}   & \copyed{88.9} \\
          \midrule
          \multicolumn{10}{c}{\textit{Multi-Task Model Fusion Methods}} \\
          Weight Averaging      & 65.3            & 63.3          & 71.4              & 73.6             & 64.2          & 52.8                   & 87.5           & 50.1            & 66.0          \\
          Fisher Merging        & \copyed{68.6}   & \copyed{69.2} & \copyed{70.7}     & \copyed{66.4}    & \copyed{72.9} & \copyed{51.1}          & \copyed{87.9}  & \copyed{59.9}   & \copyed{68.3} \\
          RegMean               & \copyed{65.3}   & \copyed{63.5} & \copyed{75.6}     & \copyed{78.6}    & \copyed{78.1} & \copyed{67.4}          & \copyed{93.7}  & \copyed{52.0}   & \copyed{71.8} \\
          Task Arithmetic       & 55.3            & 54.9          & 66.7              & 77.4             & 80.2          & 69.7                   & 97.3           & 50.1            & 69.0          \\
          Ties-Merging          & {65.0}          & {64.3}        & 74.7              & 76.8             & 81.3          & 69.4                   & 96.5           & {54.3}          & 72.8          \\
          AdaMerging (task)     & 58.3            & 53.2          & 71.8              & 80.1             & 81.6          & 84.4                   & 93.4           & 42.7            & 70.7          \\
          AdaMerging++ (task)   & \copyed{60.8}   & \copyed{56.9} & \copyed{73.1}     & \copyed{83.4}    & \copyed{87.3} & \copyed{82.4}          & \copyed{95.7}  & {\copyed{50.1}} & \copyed{73.7} \\
          AdaMerging (layer)    & 64.2            & 69.5          & 82.4              & 92.5             & 86.5          & 93.7                   & 97.7           & 61.1            & 80.9          \\
          AdaMerging++ (layer)  & \copyed{66.6}   & \copyed{68.3} & \copyed{82.2}     & \copyed{94.2}    & \copyed{89.6} & \copyed{89.0}          & \copyed{98.3}  & \copyed{60.6}   & \copyed{81.1} \\
          \midrule
          \rowcolor{mygray}
          \textbf{E-WEMoE-99\% (Ours)} & 72.4    & 70.8  &  91.0   & 95.8    & 94.9   &96.7  &  99.3  &  73.6   & 86.8 \\
          \rowcolor{mygray}
          \textbf{E-WEMoE-90\% (Ours)} & \textbf{74.3}   & 76.3  &  92.7   &  97.9   &  96.1 & 98.6 &   99.5 & \textbf{77.8}    & 89.1 \\
          \rowcolor{mygray}
          \textbf{WEMoE (Ours)} & {74.1}   & \textbf{77.4} & \textbf{93.7}     & \textbf{99.1}    & \textbf{96.2} & \textbf{98.9}          & \textbf{99.6}  & {76.4}   & \textbf{89.4} \\
          \bottomrule
        \end{tabular}
      \end{sc}
    \end{small}
  \end{center}
  \vskip -0.1in
\end{table*}
\begin{table*}[t]
  \caption{Multi-task performance when merging CLIP-ViT-L/14 models on all eight tasks.}
  \label{table:multi-task_performance_clip-vit-l-14}
   \vspace{-10pt}
  \begin{center}
    \setlength{\tabcolsep}{1.5mm}
    \small
    \begin{sc}
      \begin{tabular}{lccccccccc}
        \toprule
        \textbf{Method}       & \textbf{SUN397} & \textbf{Cars} & \textbf{RESISC45} & \textbf{EuroSAT} & \textbf{SVHN} & \textbf{GTSRB}  & \textbf{MNIST} & \textbf{DTD}           & \textbf{Avg.} \\
        \midrule
        Pre-trained           & 68.2            & 77.9          & 71.3              & 61.3             & 58.4          & 50.6            & 76.4           & 55.4                   & 64.9          \\
        Individual            & \textbf{82.3}   & \textbf{92.4} & \textbf{97.4}     & \textbf{99.9}    & \textbf{98.1} & \textbf{99.2}   & \textbf{99.7}  & 84.1                   & \textbf{94.1} \\
        Traditional MTL       & \copyed{80.8}   & \copyed{90.6} & \copyed{96.3}     & \copyed{96.3}    & \copyed{97.6} & \copyed{99.1}   & \copyed{99.6}  & \textbf{\copyed{84.4}} & \copyed{93.5} \\
        \midrule
        \multicolumn{10}{c}{\textit{Multi-Task Model Fusion Methods}}                                                                                                                              \\
        Weight Averaging      & 72.1            & 81.6          & 82.6              & 91.4             & 78.2          & 70.6            & 97.0           & 62.8                   & 79.5          \\
        Fisher Merging        & \copyed{69.2}   & \copyed{88.6} & \copyed{87.5}     & \copyed{93.5}    & \copyed{80.6} & \copyed{74.8}   & \copyed{93.3}  & \copyed{70.0}          & \copyed{82.2} \\
        RegMean               & \copyed{73.3}   & \copyed{81.8} & \copyed{86.1}     & \copyed{97.0}    & \copyed{88.0} & \copyed{84.2}   & \copyed{98.5}  & \copyed{60.8}          & \copyed{83.7} \\
        Task Arithmetic       & 74.1            & 82.1          & 86.7              & 92.6             & 87.9          & 86.8            & 98.9           & 65.6                   & 84.4          \\
        Ties-Merging          & 75.0            & 84.5          & 88.0              & 94.3             & 85.7          & 82.1            & 98.7           & 67.7                   & 84.5          \\
        AdaMerging (layer)    & \copyed{79.0}   & \copyed{90.3} & \copyed{90.8}     & \copyed{96.2}    & \copyed{93.4} & {\copyed{98.0}} & \copyed{99.0}  & {\copyed{79.9}}        & \copyed{90.8} \\
        AdaMerging++ (layer)  & {\copyed{79.4}} & \copyed{90.3} & \copyed{91.6}     & {\copyed{97.4}}  & \copyed{93.4} & \copyed{97.5}   & \copyed{99.0}  & \copyed{79.2}          & \copyed{91.0}
        \\
       \midrule
       \rowcolor{mygray}
        \textbf{E-WEMoE-99\% (Ours)} & 80.7    & 90.7   &  95.1   & 97.2    & 96.7  & 98.5 &   99.2 &   84.0  & 92.8  \\
        \rowcolor{mygray}
        \textbf{E-WEMoE-90\% (Ours)} &  \textbf{81.5}   & 92.0  & \textbf{96.0}    & 97.8    & \textbf{97.7}  & 99.1 &   99.5 & \textbf{84.1}    &  93.5 \\
        \rowcolor{mygray}
        \textbf{WEMoE (Ours)} & {81.4}   & \textbf{92.6} & {95.4}     & \textbf{99.4}    & \textbf{97.7} & \textbf{99.3}   & \textbf{99.7}  & {83.7}          & \textbf{93.6} \\
        \bottomrule
      \end{tabular}
    \end{sc}
  \end{center}
  \vskip -0.2in
\end{table*}

\subsection{Experimental Setup}
\label{subsec:setup}

In this paper, we use CLIP~\citep{radfordLearningTransferableVisual2021} (e.g., CLIP-ViT-B/32, CLIP-ViT-B/16 and CLIP-ViT-L/14) as the pre-trained model. CLIP is trained on a large-scale dataset consisting of image-text pairs and is capable of performing open-vocabulary image classification effectively. Following the approach of Task Arithmeitc~\cite{ilharcoEditingModelsTask2023}, we fine-tune the model on eight image classification tasks: SUN397 \citep{xiao_sun_2010}, Stanford Cars \citep{krause_3d_2013}, RESISC45 \citep{cheng_remote_2017}, EuroSAT \citep{helber2018introducing}, SVHN \citep{netzer_reading_2021}, GTSRB \citep{stallkamp_man_2012}, MNIST \citep{lecunGradientbasedLearningApplied1998}, and DTD \citep{cimpoi_describing_2014}. We then report the Top-1 accuracy for each task, as well as the average accuracy (AVG.) across all tasks.

We compare the proposed WEMoE and E-WEMoE with several popular model merging methods, including Fisher Merging~\citep{matenaMergingModelsFisherWeighted2022}, RegMean~\citep{jinDatalessKnowledgeFusion2023}, Task Arithmetic~\citep{ilharcoEditingModelsTask2023},  Ties-Merging~\citep{yadavResolvingInterferenceWhen2023}, and AdaMerging/Adamerging++~\citep{yangAdaMergingAdaptiveModel2023}. Additionally, we report the performance of three baseline approaches: individual models, the traditional MTL model, and the pre-trained model, as a reference. For all methods, unless otherwise specified, we follow the configuration in AdaMerging~\citep{yangAdaMergingAdaptiveModel2023} and set the initial merging coefficient $\lambda$ of the task vector to 0.3.

\subsection{Multi-Task Model Merging Performance}
\label{subsec:performance}

Table~\ref{table:multi-task_performance_clip-vit-b-32} and Table~\ref{table:multi-task_performance_clip-vit-l-14} provide a performance comparison of model merging methods applied to ViT-B/32 and ViT-L/14. Due to space constraints, the results for ViT-B/16 (Table~\ref{table:multi-task_performance_clip-vit-b-16}) are included in the Appendix. 

The following observations can be made from these tables: 
(1) The pre-trained model shows the poorest performance, as it lacks information specific to the downstream tasks. In contrast, the expert models, either independently fine-tuned or trained using traditional multi-task learning, perform exceptionally well due to their full incorporation of downstream task information. 
(2) Larger architectures, such as ViT-L/14, tend to be easier to merge than smaller ones like ViT-B/32. This is because larger models often contain more redundant parameters and exhibit stronger cross-task generalization capabilities.
(3) Advanced merging methods, such as Ties-Merging and AdaMerging, outperform basic approaches like weight averaging or task arithmetic by effectively addressing parameter conflicts and merging models in a more fine-grained way. 
(4) The proposed WEMoE method outperforms other merging techniques by dynamically combining cross-task shared knowledge with task-specific knowledge, allowing it to better adapt to input data. Notably, WEMoE outperforms the traditional MTL baseline by 0.5\% and 0.1\% on average. 
(5) E-WEMoE further showcases the advantages of both high performance and efficiency by retaining WEMoE's dynamic merging capabilities while reducing the number of trainable and total parameters through MLP task vector sparsification and cross-layer sharing. For instance, on ViT-L/14, E-WEMoE-90\% achieves a performance close to WEMoE (93.5 vs. 93.6) while reducing the total parameters by 4 times and trainable parameters by 24 times (as depicted in Figure \ref{fig:num_pram}).
In summary, the WEMoE and E-WEMoE proposed in this paper show strong merging performance.

\begin{table*}[t]
  \caption{Generalization results on two unseen tasks when merging ViT-B/32 models on six tasks.}
  \label{table:generalization_results_clip-vit-b-32}
   \vspace{-10pt}
  \begin{center}
    \setlength{\tabcolsep}{1.5mm}
    \begin{small}
      \begin{sc}
        \begin{tabular}{l|ccccccc|ccc}
          \toprule
          \multirow{2}{*}{\textbf{Method}} & \multicolumn{7}{c|}{\textbf{Seen Tasks}} & \multicolumn{3}{c}{\textbf{Unseen Tasks}}                                                                                                                                          \\
                                           & SUN397                                   & Cars                                      & RESISC45      & DTD           & SVHN          & GTSRB         & \textbf{Avg.} & MNIST         & EuroSAT                & \textbf{Avg.} \\
          \midrule
          Task Arithmetic                  & 63.4                                     & 62.3                                      & 75.3          & 57.8          & 84.7          & 80.4          & 70.7          & 77.3          & 45.6                   & 61.4          \\
          Ties-Merging                     & 67.8                                     & 66.2                                      & 77.0          & 56.2          & 77.2          & 71.0          & 69.2          & 75.9          & 43.1                   & 59.5          \\
          AdaMerging                       & \copyed{65.2}                            & \copyed{65.9}                             & \copyed{88.5} & \copyed{61.1} & \copyed{92.2} & \copyed{91.5} & \copyed{77.4} & \copyed{84.0} & \textbf{\copyed{56.1}} & \copyed{70.0} \\
          \midrule
           \rowcolor{mygray}
          \textbf{E-WEMoE (2-layer)}        
          &\textbf{74.6} &77.0   & 93.3  &\textbf{76.7}  & 96.4   &98.4 &86.0 &73.8 &49.7 & {61.7} \\
          \rowcolor{mygray}
          \textbf{WEMoE (0-layer)}         & 63.8                                     & 61.2                                      & 78.4          & 56.4          & 89.1          & 92.2          & 73.5          & 78.6          & 49.7                   & 64.2          \\
          \rowcolor{mygray}
          \textbf{WEMoE (2-layer)}         & {74.4}                            & \textbf{78.3}                             & \textbf{94.8} & {75.6} & \textbf{96.8} & \textbf{99.0} & \textbf{86.5} & \textbf{86.3} & 55.9                   & \textbf{71.1} \\
          \bottomrule
        \end{tabular}
      \end{sc}
    \end{small}
  \end{center}
  \vskip -0.1in
\end{table*}

\begin{table*}[t]
  \caption{Ablations of the test data distribution on ViT-B/32 (for all methods, $\lambda=0.3$).}
  \label{table:abalation_data_distribution_vit_b_32}
   \vspace{-10pt}
  \begin{center}
    \setlength{\tabcolsep}{1.5mm}
    \fontsize{8}{9}\selectfont
    \begin{sc}
      \begin{tabular}{l|ccccc|ccccc}
        \toprule
        \textbf{Method}          & \textbf{Cars}                                           & \textbf{EuroSAT}                                          & \textbf{RESISC45} & \textbf{GTSRB} & \textbf{Avg.} & \textbf{Cars} & \textbf{EuroSAT} & \textbf{RESISC45} & \textbf{GTSRB} & \textbf{Avg.} \\
        \midrule
                                 & \multicolumn{5}{c|}{{Clean Test Set}}                   & \multicolumn{5}{c}{{Corrupted Test Set (Motion Blur)}}                                                                                                                                                 \\
        Task Arithmetic          & 66.9                                                    & 94.7                                                      & 82.6              & 75.1           & 79.8          & 65.3          & 68.1             & 80.0              & 64.2           & 69.4          \\
        Ties-Merging             & 67.5                                                    & 83.7                                                      & 79.8              & 65.3           & 74.1          & 65.6          & 57.5             & 77.5              & 55.4           & 64.0          \\
        AdaMerging               & 73.7                                                    & 96.1                                                      & 85.8              & 96.3           & 88.0          & 71.2          & 74.6             & 82.7              & 94.1           & 80.6          \\
        \rowcolor{mygray}
        \textbf{E-WEMoE (2-layer)} &77.5 &98.4 &93.7 &98.7 &92.1  
        &76.1 &78.3 &93.6 &97.5 &86.4 
       \\
        \rowcolor{mygray}
        \textbf{WEMoE (2-layer)} & \textbf{78.8}                                           & \textbf{99.5}                                             & \textbf{95.4}     & \textbf{99.1}  & \textbf{93.2} & \textbf{78.1} & \textbf{79.7}    & \textbf{94.6}     & \textbf{97.8}  & \textbf{87.6} \\
        \midrule
        & \multicolumn{5}{c|}{Corrupted Test Set (Impluse Noise)} & \multicolumn{5}{c}{Corrupted Test Set (Gaussian Noise)}                                                                                                                                                \\
        Task Arithmetic          & 62.1                                                    & \textbf{49.1}                                             & 72.7              & 40.4           & 56.1          & 63.6          & \textbf{55.4}    & 75.9              & 49.4           & 61.1          \\
        Ties-Merging             & 62.1                                                    & 42.0                                                      & 70.4              & 34.9           & 52.3          & 64.1          & 50.3             & 74.5              & 39.8           & 57.2          \\
        AdaMerging               & 67.2                                                    & 30.8                                                      & 75.9              & 77.5           & 62.8          & 69.9          & 41.2             & 80.6              & 76.0           & 66.9          \\
        \rowcolor{mygray}
        \textbf{E-WEMoE (2-layer)} &73.7 &12.3 &90.7 &\textbf{91.9} &67.2 &75.5 &15.2 &91.6 &\textbf{88.7} &67.8 \\
        \rowcolor{mygray}
        \textbf{WEMoE (2-layer)} & \textbf{74.7}                                           & 11.6                                                      & \textbf{91.4}     & {91.7}  & \textbf{67.3} & \textbf{76.8} & 29.7             & \textbf{93.2}     & {78.2}  & \textbf{69.5} \\
        \midrule
                                 & \multicolumn{5}{c|}{Corrupted Test Set (Pixelate)}      & \multicolumn{5}{c}{Corrupted Test Set (Spatter)}                                                                                                                                                       \\
        Task Arithmetic          & 2.8                                                     & 41.5                                                      & 22.8              & 66.6           & 33.4          & 63.3          & \textbf{60.1}    & 73.9              & 54.3           & 62.9          \\
        Ties-Merging             & \textbf{4.1}                                            & 40.8                                                      & 20.6              & 57.1           & 30.6          & 64.4          & 50.8             & 71.4              & 44.3           & 57.8          \\
        AdaMerging               & 2.5                                                     & \textbf{53.8}                                             & 22.4              & 90.6           & \textbf{42.3} & 69.9          & 43.6             & 75.4              & 89.4           & 69.6          \\
        \rowcolor{mygray}
        \textbf{E-WEMoE (2-layer)} &0.5 &23.0 &4.4 &96.6 &31.1 &73.0 &22.8 &90.6 &95.9 &70.6 \\
        \rowcolor{mygray}
        \rowcolor{mygray}
        \textbf{WEMoE (2-layer)} & 0.4                                                     & 9.6                                                       & 2.2               & \textbf{97.0}  & 27.3          & \textbf{76.2} & 28.2             & \textbf{91.2}     & \textbf{96.0}  & \textbf{72.9} \\
        \midrule
                                 & \multicolumn{5}{c|}{Corrupted Test Set (Contrast)}      & \multicolumn{5}{c}{Corrupted Test Set (JPEG Compression)}                                                                                                                                              \\
        Task Arithmetic          & 66.0                                                    & 62.9                                                      & 75.9              & 70.6           & 68.9          & 66.5          & 72.3             & 82.2              & 60.0           & 70.3          \\
        Ties-Merging             & 66.8                                                    & 53.4                                                      & 75.9              & 61.5           & 64.4          & 67.5          & 60.4             & 80.0              & 50.1           & 64.5          \\
        AdaMerging               & 71.7                                                    & 69.8                                                      & 79.3              & 95.1           & {79.0}        & {70.9}        & 75.8             & 83.6              & 90.1           & 80.1          \\
        \rowcolor{mygray}
        \textbf{E-WEMoE (2-layer)} &77.1 &70.1 &93.2 &\textbf{98.5} &84.7 
        &76.8 &\textbf{85.8} &93.6 &94.9 &\textbf{87.8} \\
        \rowcolor{mygray}
        \textbf{WEMoE (2-layer)} & \textbf{77.7}                                           & \textbf{77.4}                                             & \textbf{93.9}     & \textbf{98.5}  & \textbf{86.9} & \textbf{78.2} & {80.7}    & \textbf{95.1}     & \textbf{96.2}  & {87.6} \\
        \bottomrule
      \end{tabular}
    \end{sc}
  \end{center}
  \vskip -0.2in
\end{table*}

\subsection{Generalization and Robustness Evaluation}
\label{eq:generalization_robustness}

The primary objective of model merging is to preserve performance on the downstream tasks it has been trained on (as demonstrated in Section \ref{subsec:performance}). However, in real-world applications, once the model is deployed, it often faces unseen tasks or experiences significant distribution shifts between the test and training data. Therefore, evaluating the generalization and robustness of the merged model becomes crucial.

\textbf{Generalization}.
To assess the generalization capability of our approach, we merged models fine-tuned on six tasks (SUN397, CARS, RESISC45, DTD, SVHN, and GTSRB) and evaluated them on two unseen tasks (MNIST and EUROSAT). As shown in Table~\ref{table:generalization_results_clip-vit-b-32}, E-WEMoE achieved an 86.0 performance on the seen tasks, outperforming AdaMerging, Ties-Merging, and Task Arithmetic, while WEMoE reached an even higher performance of 86.5. Additionally, on the unseen tasks, WEMoE demonstrated superior performance compared to other baselines, indicating strong generalization to new tasks.

\textbf{Robustness}.
To evaluate the robustness of the merging algorithm, we generated distorted images from the clean test set following the approach in \citep{hendrycksBenchmarkingNeuralNetwork2019}. Specifically, we adopted the settings from AdaMerging~\cite{yangAdaMergingAdaptiveModel2023}, using seven types of distortions: motion blur, impulse noise, gaussian noise, pixelate, spatter, contrast, and JPEG compression. Examples of distorted images on the Cars dataset are shown in Figure \ref{fig:distorted_images} in the Appendix. 

We evaluated the performance of the merged model on both clean and distorted images, presenting the results for ViT-B/32 in this section, with ViT-B/16 results provided in the Appendix (Table \ref{table:abalation_data_distribution_vit_b_16}) due to space constraints. On the ``clean" dataset, WEMoE and E-WEMoE clearly demonstrate the best performance. Moreover, these methods consistently show significant advantages across various corrupted datasets. For instance, under ``motion blur", E-WEMoE achieved a performance of 86.4, and WEMoE reached 87.6, outperforming the best baseline, AdaMerging, which scored 80.6. An exception occurs with the ``pixelate" distribution, where WEMoE and E-WEMoE do not perform optimally due to the severe image degradation, as illustrated in Figure \ref{fig:distorted_images}(e), making the images difficult to recognize. However, under the other six types of distortions, both WEMoE and E-WEMoE consistently achieve the best results, demonstrating the robustness of our approach.

\subsection{Analysis}
\label{subsec:analysis}
\textbf{Ablations of the Up-scaling Strategy}.
In the methods section, we primarily focus on upscaling the MLP module. Here, we explore additional variants: (1) \textit{Entire Transformer Block}, which integrates the entire block into the MoE structure; (2) \textit{Att + MLP}, which constructs the Attention module and MLP as separate MoE structures; and (3) \textit{MLP Only}, our default method that upgrades the MLP to an MoE structure. The results, presented in Table~\ref{table:upscaling_strategies}, indicate that upcycling the entire block to MoE is not the optimal choice. Additionally, when the Attention and MLP weights are upgraded separately, there is a significant performance improvement across most tasks, surpassing the other two variants. However, this approach requires maintaining a larger number of parameters compared to merely upscaling the MLP. Therefore, considering the marginal performance improvement against the increased computational load and memory usage, we conclude that upgrading only the MLP is a preferable choice.
\begin{table}[h]
\vskip -0.05in
  \caption{Comparison of different up-scaling strategies.
  }
   \vspace{-5pt}
  \label{table:upscaling_strategies}
  \centering
  \small
  \begin{tabular}{lccccccccc}
    \toprule
    \textbf{Method (1-Layer Router)} & \textbf{Avg.} \\
    \midrule
    Entire Transformer Block    & 86.4          \\
    Att + MLP (Separately) & \textbf{89.2} \\
    MLP Only    & 88.3          \\
    \bottomrule
  \end{tabular}
  \vskip -0.05in
\end{table}

\textbf{Routing Activation Analysis}.
In this part, we provide a brief analysis of the output weights from the routing mechanism, with a more detailed discussion available in the appendix. Using ViT-B/32 as an example, we examine the routing output weights across different Transformer blocks ($\small l=\{0, 3, 6, 9, 11\}$) for each task. As shown in Figure \ref{fig:router_analysis} and Figure \ref{fig:ewemoe_router_analysis}, we observe that both WEMoE and E-WEMoE consistently assign higher weights to the source tasks most related to the input samples. This demonstrates that our routing mechanism effectively identifies the top-performing experts based on the input features. Even in E-WEMoE, where cross-layer shared routing is employed, the routers allocate greater weights to the most relevant MLP task vectors, aligning with our intuitive expectations and design objectives.

\section{Related Work}
\label{section:related_work}

Weight interpolation has proven to be a simple yet effective strategy for scalable model merging, requiring minimal computational overhead~\citep{izmailovAveragingWeightsLeads2019,matenaMergingModelsFisherWeighted2022,wortsmanModelSoupsAveraging2022,kaddourStopWastingMy2022,ilharcoEditingModelsTask2023,yadavResolvingInterferenceWhen2023,yangAdaMergingAdaptiveModel2023,wuPiTuningTransferring2023,RepresentationSurgery_ICML_2024}. A comprehensive discussion on model merging is available in \cite{Survery_ModelMerging_2024}. 
However, achieving successful model merging remains challenging due to the potential for task interference. 

Several advanced methods have been proposed to address performance degradation in model merging. 
(1) Some approaches emphasize that the choice of merging coefficients is crucial to final performance, leading to the development of various strategies for optimizing these weights~\cite{matenaMergingModelsFisherWeighted2022,jinDatalessKnowledgeFusion2023,yangAdaMergingAdaptiveModel2023}. However, these methods typically use static merging weights, whereas this paper introduces a dynamic approach to weight assignment. 
(2) Some works argue that neural networks are over-parameterized, with many neurons being ineffective for downstream tasks and potentially causing task conflicts and interference; thus, they propose merging in a subspace~\cite{yadavResolvingInterferenceWhen2023,guodong24neurips,tangConcreteSubspaceLearning2023,yu2024language,wang2024localizing,huang2024emr}. In contrast, this paper differs by applying sparsification specifically to the MLP modules rather than the entire parameter space. Moreover, while previous methods perform static merging, this paper combines sparse task vectors through dynamic merging.
We discuss more model merging works in the Appendix \ref{section:related_work_appendix}.

\section{Conclusion and Future Work}
\label{section:conclusion}

This paper introduces WEMoE, a novel approach for merging multiple expert models of Transformer-based vision models to create a unified MTL model. WEMoE enhances key modules in the Transformer block with a mixture-of-experts (MoE) structure while statically merging the non-key modules. To address the additional storage costs and increased number of trainable parameters associated with WEMoE, we propose an efficient variant, E-WEMoE, which reduces parameter storage by pruning non-essential elements in the key modules and utilizes cross-layer shared routing to minimize the number of trainable parameters. Extensive experiments across three architectures and eight tasks demonstrate the method's strong performance, generalization capabilities, and robustness.

Future work will explore extending our approach to more application scenarios, such as merging Transformer models from different components or integrating parameter-efficient fine-tuning techniques like Adapters\citep{houlsbyParameterEfficientTransferLearning2019} or LoRA~\citep{huLoRALowRankAdaptation2021} to further enhance parameter efficiency.

{
\small
\bibliographystyle{IEEEtran}
\bibliography{main}
}

\clearpage


\clearpage

\noindent
\textbf{Appendix Overview}. 
The appendix is organized into several sections, each providing additional insights and details related to different aspects of the main work.

{
\hypersetup{linkcolor=black}
\startcontents[sections] 
\printcontents[sections]{}{}{\setcounter{tocdepth}{2}} 
\vskip 0.2in
\hrule
}

\section{Related Work}
\label{section:related_work_appendix}

In this section, we review recent advancements in multi-task model merging and mixture of experts (MoE) methods. 

\subsection{Multi-Task Model Merging}
\label{subsection:multi_task_model_fusion}

Model merging offers an efficient approach to integrating the strengths of multiple expert models into a single model~\citep{izmailovAveragingWeightsLeads2019,matenaMergingModelsFisherWeighted2022,wortsmanModelSoupsAveraging2022,kaddourStopWastingMy2022,ilharcoEditingModelsTask2023,yadavResolvingInterferenceWhen2023,yangAdaMergingAdaptiveModel2023,tangParameterEfficientMultitask2023,wuPiTuningTransferring2023,RepresentationSurgery_ICML_2024,yang2024surgeryv2,ortiz2023task,singh2020model}. By performing weight interpolation directly at the parameter level, it serves as a powerful alternative to traditional multi-task learning~\cite{yadav2024survey,Survery_ModelMerging_2024}.

By exploring the loss landscape, researchers have uncovered the phenomenon of mode connectivity \citep{danielfreemanTopologyGeometryHalfrectified2017,nagarajanUniformConvergenceMay2019}, which suggests that different solutions can be linked by paths in the parameter space that maintain low loss values. This insight has facilitated model fusion \citep{draxlerEssentiallyNoBarriers2019,frankleLinearModeConnectivity2020,entezariRolePermutationInvariance2022}. 
This technique involves aligning and interpolating corresponding components across models to minimize discrepancies \citep{liConvergentLearningDifferent2016,tatroOptimizingModeConnectivity2020}. Common methods include matching activations or weights \citep{georgestoicaZipItMergingModels2023}, applying graph matching for channel alignment \citep{liuDeepNeuralNetwork2022a}, and leveraging the principle of permutation invariance \citep{ainsworthGitReBasinMerging2023}.
In general, mode connectivity ensures the feasibility of model merging, while parameter alignment facilitates more effective merging by providing a crucial prerequisite. This approach complements research efforts to mitigate task interference during model merging, representing an orthogonal yet synergistic direction.

This paper focuses on developing improved model merging techniques to reduce task interference. Existing approaches in this area include weighted merging, subspace merging, and others. 
(1) \textit{Weighted-based merging methods} posit that different models should carry varying levels of importance during the merging process, with the importance dimension potentially refined to the task, layer, or parameter level~\cite{matenaMergingModelsFisherWeighted2022,wortsmanModelSoupsAveraging2022,jinDatalessKnowledgeFusion2023,yangAdaMergingAdaptiveModel2023,zhou2024metagpt,akiba2024evolutionary,daheim2023model}. The core objective is to identify the optimal merging coefficients based on predefined rules or learnable methods.
(2) \textit{Subspace-based Merging methods} aim to eliminate unimportant neurons in the models being merged, thereby alleviating interference between tasks~\cite{yadavResolvingInterferenceWhen2023,guodong24neurips}. The core of these approaches is to identify a small set of neurons that are most critical for downstream tasks~\cite{tangConcreteSubspaceLearning2023,yu2024language,wang2024localizing,huang2024emr,zimmer2023sparse,he2024localize,deep2024della,davari2023model,panigrahi2023task}.
However, these methods rely on \textit{static} merging strategies, meaning that once the merging is completed, the model's structure and parameters are fixed, and all input samples are processed using the same merging weights. This approach is suboptimal as it restricts the model's adaptability to different samples.

Recently, a small number of works have attempted to dynamically merge expert modules~\cite{lu2024twin,muqeeth2023soft,komatsuzaki2022sparse,ye2023merging,muqeeth2024learning}.  However, these works either concentrate on increasing the model scale during the training phase~\cite{komatsuzaki2022sparse} or focus on zero-shot generalization~\cite{muqeeth2024learning} rather than multi-task model merging like us. Additionally, some remaining works~\cite{lu2024twin,muqeeth2023soft,ye2023merging} merge dense expert modules, which incur higher memory costs compared to merging sparse MLP task vectors. 
Overall, our approach combines both high efficiency and effectiveness.

\subsection{Mixture of Experts}
\label{subsection:mixture_of_experts}

The mixture of experts (MoE) model, first proposed by \cite{jacobsAdaptiveMixturesLocal1991}, is a machine learning technique that involves training multiple models, each specialized to handle a different part of the input space. Over the years, MoE has gained considerable attention \citep{jiangMixtralExperts2024,daiDeepSeekMoEUltimateExpert,zhouMixtureofExpertsExpertChoice2022}. Many innovations have focused on developing more efficient router designs~\cite{fedusSwitchTransformersScaling2022,lewisBASELayersSimplifying2021}. For instance, the Switch Transformer~\citep{fedusSwitchTransformersScaling2022} simplifies the selection process by choosing only the top expert per token, showcasing superior scalability compared to prior approaches. Base Layers~\citep{lewisBASELayersSimplifying2021} introduces a linear assignment that optimizes token-expert affinities, ensuring an equal distribution of tokens among experts. In addition to methods that involve routers selecting experts, there are alternative approaches such as allowing each expert to choose tokens~\citep{zhouMixtureofExpertsExpertChoice2022}.
Since this paper does not focus on the design of MoE, more comprehensive work can be found in the survey paper \citep{fedusReviewSparseExpert2022}.

\section{Loss Landscape Visualization}
\label{appendix:loss}

\begin{figure*}[h]
    \centering
    \begin{minipage}{0.32\textwidth}
        \centering
        \includegraphics[width=0.85\linewidth]{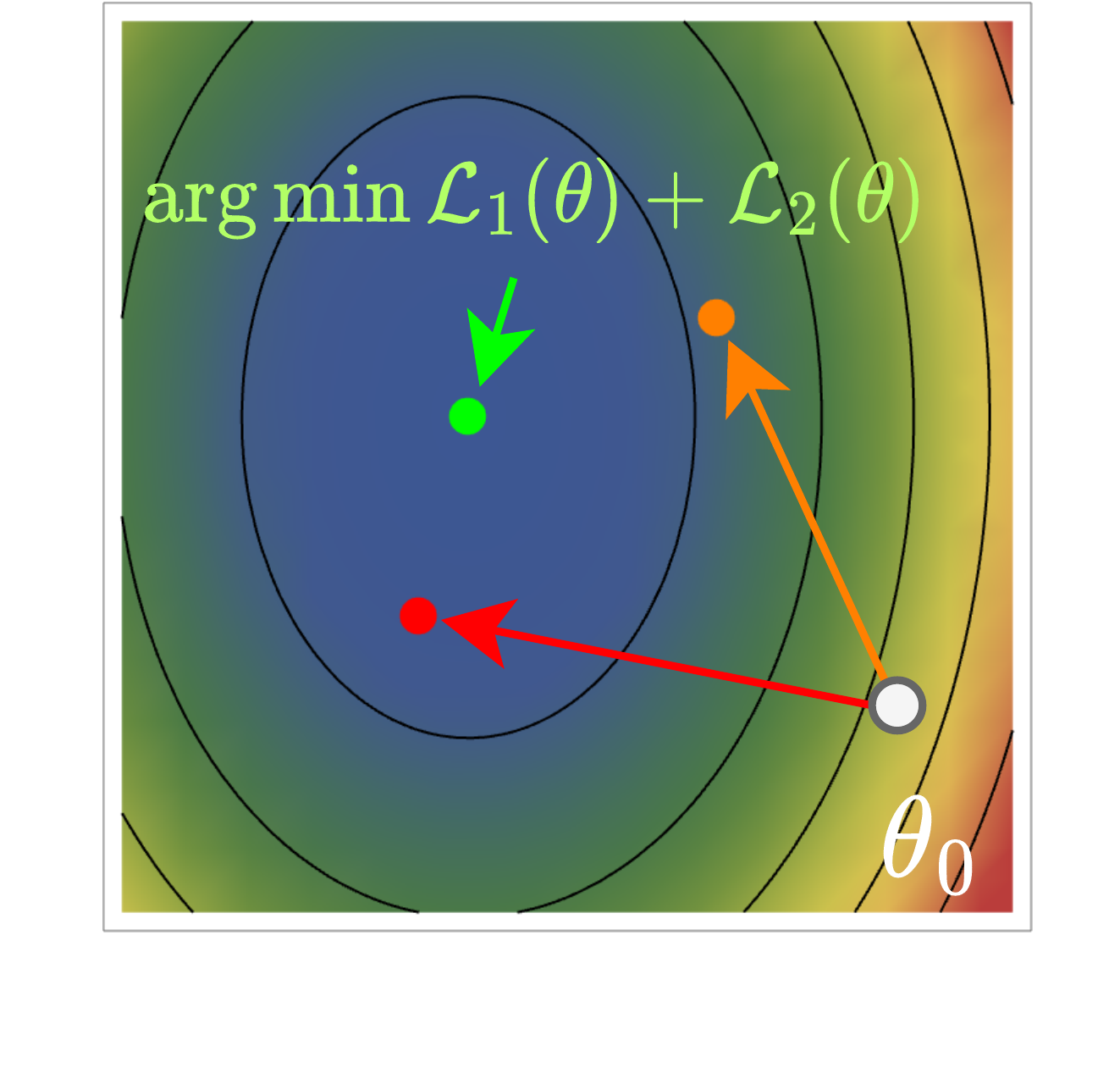}
        \vspace{-20pt}
       \begin{center}
           (a) $\mathcal{L}_1 + \mathcal{L}_2$
       \end{center}
    \end{minipage}
    \hfill
    \begin{minipage}{0.32\textwidth}
        \centering
       \includegraphics[width=\linewidth]{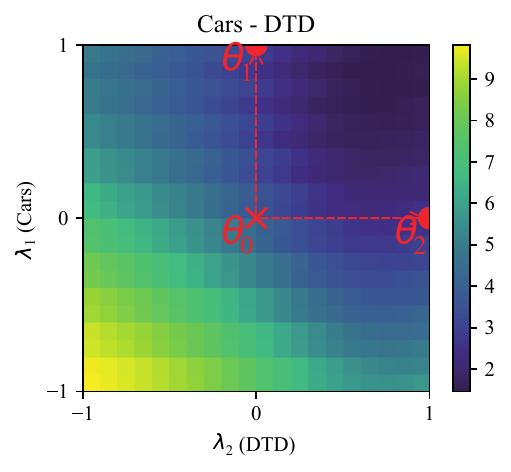}
        \vspace{-20pt}
       \begin{center}
           (b) Cars-DTD
       \end{center}
    \end{minipage}
    \hfill
    \begin{minipage}{0.32\textwidth}
        \centering
        \includegraphics[width=\linewidth]{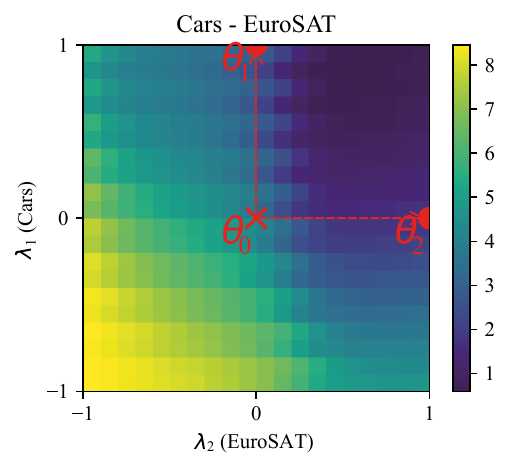}
        \vspace{-20pt}
       \begin{center}
           (c) Cars-EuroSAT
       \end{center}
    \end{minipage}
    
    \vspace{0.1cm} 
    
    \begin{minipage}{0.32\textwidth}
        \centering
       \includegraphics[width=\linewidth]{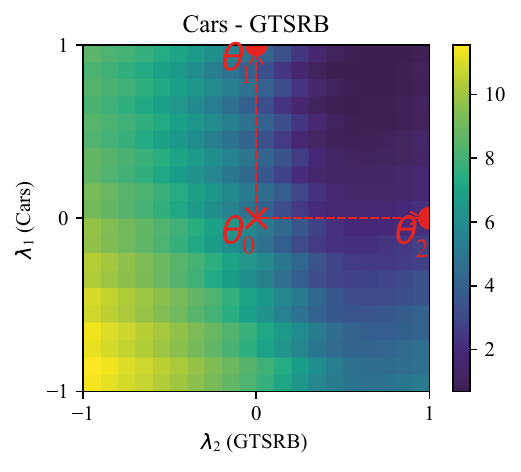}
        \vspace{-20pt}
       \begin{center}
           (d) Cars-GTSRB
       \end{center}
    \end{minipage}
    \hfill
    \begin{minipage}{0.32\textwidth}
        \centering
       \includegraphics[width=\linewidth]{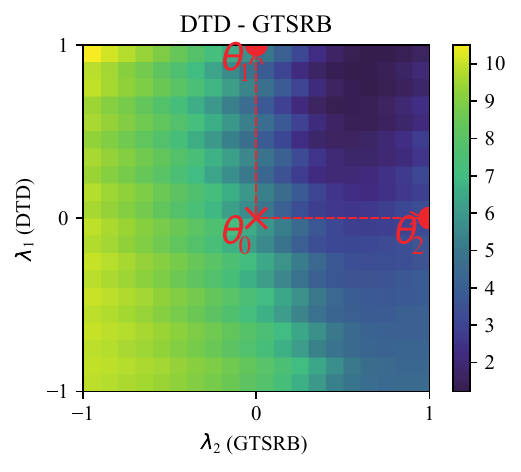}
        \vspace{-20pt}
       \begin{center}
           (e) DTD-GTSRB
       \end{center}
    \end{minipage}
    \hfill
    \begin{minipage}{0.32\textwidth}
        \centering
        \includegraphics[width=\linewidth]{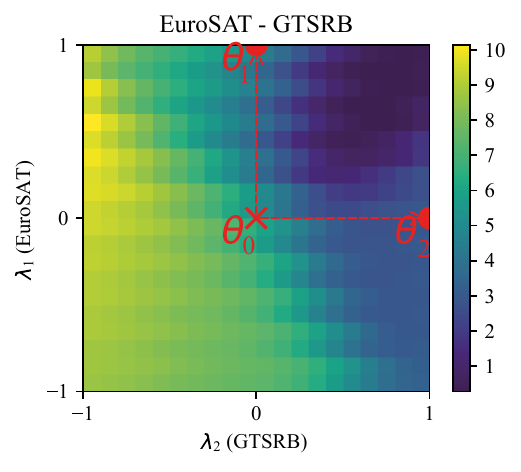}
        \vspace{-20pt}
       \begin{center}
           (f) EuroSAT-GTSRB
       \end{center}
    \end{minipage}
    \caption{
    Visualization of (a) the joint loss $\mathcal{L}_1 + \mathcal{L}_2$ and (b-f) five task pairs for ViT-B/32 in the loss landscape.
  }
  \label{fig:loss_landscapes_examples}
\vspace{-10pt}
\end{figure*}

In Section~\ref{subsection:revisiting_model_merge}, we emphasized that static solutions for multi-task model merging have limitations, making it challenging to find a set of merging parameters that are optimal for all tasks. To further illustrate this point, we present joint loss landscapes for five sets of real task pairs using ViT-B/32. 

These loss landscapes are obtained by interpolating between the pretrained model and the expert models for tasks 1 and 2, formally expressed as $\theta = \theta_0 + \lambda_1 \tau_1 + \lambda_2 \tau_2$, where $\theta_0$ represents the pre-trained weights, $\tau_i = \theta_i - \theta_0$ is the vector constructed from the two tasks, and $\lambda_i$ ranges from -1 to 1. As shown in Figure~\ref{fig:loss_landscapes_examples}, the expert models fine-tuned from the same pre-trained model occupy the same loss basin on the joint task; however, there is no set of interpolated weights $\lambda_1$ and $\lambda_2$ that simultaneously minimizes the loss for both tasks.

\section{Multi-Task Model Merging}
\label{appendix:multi_task_model_fusion}

\begin{table}[h]
  \caption{Comparison of different methods and their data access requirements.}
  \label{table:method_comparison}
  \centering
  \small
\resizebox{\linewidth}{!}{  
  \begin{tabular}{lccc}
    \midrule
    \textbf{Method}                                              & \textbf{Labeled} & \textbf{Labeled} & \textbf{Test time} \\
     & \textbf{Tasks data} & \textbf{Validation data} & \textbf{Adaptation} \\
    \midrule
    Fisher Merging~\citep{matenaMergingModelsFisherWeighted2022} & Yes                         & No                                 & No                            \\
    RegMean~\citep{jinDatalessKnowledgeFusion2023}               & Yes                         & No                                 & No                            \\
    Task Arithmetic~\citep{ilharcoEditingModelsTask2023}         & No                          & Yes                                & No                            \\
    Ties-Merging~\citep{yadavResolvingInterferenceWhen2023}      & No                          & Yes                                & No                            \\
    AdaMerging~\citep{yangAdaMergingAdaptiveModel2023}           & No                          & No                                 & Yes                           \\
    \rowcolor{mygray}
    \textbf{E-WEMoE (Ours)}                                                & No                          & No                                 & Yes                           \\
    \rowcolor{mygray}
    \textbf{WEMoE (Ours)}                                                & No                          & No                                 & Yes                           \\
    \bottomrule
  \end{tabular}
  }
  \vspace{-10pt}
\end{table}

\begin{table*}[t]
  \caption{Multi-task performance when merging CLIP-ViT-B/16 models on all eight tasks.}
  \label{table:multi-task_performance_clip-vit-b-16}
  \begin{center}
    \setlength{\tabcolsep}{1.5mm}
    \small
    \begin{sc}
      \begin{tabular}{lccccccccc}
        \toprule
        \textbf{Method}       & \textbf{SUN397} & \textbf{Cars} & \textbf{RESISC45} & \textbf{EuroSAT} & \textbf{SVHN} & \textbf{GTSRB}  & \textbf{MNIST} & \textbf{DTD}           & \textbf{Avg.} \\
        \midrule
        Pre-trained           & 65.5         &64.6         &66.3              &  54.1     &    51.9 &  43.4            &51.7       &44.9                    &55.3      \\
        Individual            &   \textbf{78.9}       &   \textbf{85.9}      &  \textbf{96.6}            &  \textbf{99.0}     & \textbf{97.6}   &         \textbf{99.0}     &    \textbf{99.7}   &    \textbf{82.3}               & \textbf{92.3}     \\
        \midrule
        \multicolumn{10}{c}{\textit{Multi-Task Model Fusion Methods}}                                                                                                                              \\
        Weight Averaging      &68.7          & 69.0        & 75.0             &83.2       & 74.9   & 62.5             & 93.7      & 51.1                   &  72.3    \\
        Fisher Merging       &  68.8        & 71.0        &  80.9            & 93.8       & 76.8   &  74.4            & 88.6      & 52.0                   & 75.8     \\
        RegMean               & 71.2         &76.4         &   86.6           &  95.1     & 94.0   &  86.5            & 98.5      & 63.9                    & 84.0     \\
        Task Arithmetic      & 65.9         &   68.3      &75.4              & 84.5      & 88.8   &   81.9           & 98.0      & 53.9                   &  77.1    \\
        Ties-Merging         & 70.6         &71.2         &79.8              &  87.5     &83.2    & 76.2             &    96.4   &55.4                    & 77.5     \\
        AdaMerging (layer)   & 70.6         &  79.9       & 86.5             &  93.5     & 93.7   &95.3              & 98.1      &63.0                    &    85.1  \\
       \midrule
       \rowcolor{mygray}
        \textbf{E-WEMoE-99\% (Ours)} &  76.3        &  80.7       &  94.2            & 95.6      &  96.5  &   98.1           &  99.2     &   79.8                 &  90.1    \\
        \rowcolor{mygray}
        \textbf{E-WEMoE-90\% (Ours)} &  \textbf{77.7}        &  \textbf{85.0}      &  \textbf{94.9}            & 98.2      & 97.2   &  \textbf{98.9}            &  99.5     & \textbf{81.4}                   &  \textbf{91.6}    \\
        \rowcolor{mygray}
        \textbf{WEMoE (Ours)} &77.2          & \textbf{85.0}         & 94.8             & \textbf{98.3}      &\textbf{97.3}    & \textbf{98.9}             &   \textbf{99.6}    &80.8                    &  91.5    \\
        \bottomrule
      \end{tabular}
    \end{sc}
  \end{center}
  \vskip -0.1in
\end{table*}

\subsection{Model Merging Performance}
In the multi-task performance comparison, we evaluated various methods, including traditional multi-task learning (MTL), weighted averaging, Fisher merging \citep{matenaMergingModelsFisherWeighted2022}, RegMean \citep{jinDatalessKnowledgeFusion2023}, task algorithm \citep{ilharcoEditingModelsTask2023}, ties-merging \citep{yadavResolvingInterferenceWhen2023}, and AdaMerging/AdaMerging++ (task-wise and layer-wise) \citep{yangAdaMergingAdaptiveModel2023}, along with our proposed E-WEMoE and WEMoE methods. We further detail the differences among these methods in Table \ref{table:method_comparison}. Specifically, our methods do not rely on any labeled data; instead, they require a test-time adaptation process, which is highly efficient and can be completed within a few minutes.

Tables \ref{table:multi-task_performance_clip-vit-b-32} and \ref{table:multi-task_performance_clip-vit-l-14} in the main text present a comprehensive comparison of the multi-task model merging performance for the CLIP-ViT-B/32 and CLIP-ViT-L/14 models across eight different tasks. In this section, we further discuss the performance comparison for the CLIP-ViT-B/16 model in Table~\ref{table:multi-task_performance_clip-vit-b-16}. We present the following key observations that demonstrate the effectiveness of WEMoE in multi-task model merging: 
(1) The independently fine-tuned model exhibits no task interference, resulting in the best performance. However, it requires maintaining a separate model for each task, leading to an extremely large number of parameters that need to be stored. 
(2) Various static model merging methods yield only one final model, which has a parameter count equivalent to the pre-trained model while significantly outperforming it. Still, there is a significant gap with the individual model.
(3) Our proposed WEMoE dynamic merging model achieves performance closest to that of the independently fine-tuned model. 
(4) E-WEMoE further reduces both the total number of parameters and the number of trainable parameters compared to WEMoE, while still achieving performance levels similar to those of WEMoE.

\subsection{Robustness to Out-of-Distribution Data}
\begin{figure}[htbp]
    \centering
    \begin{minipage}{0.11\textwidth}
        \centering
        \includegraphics[width=\linewidth]{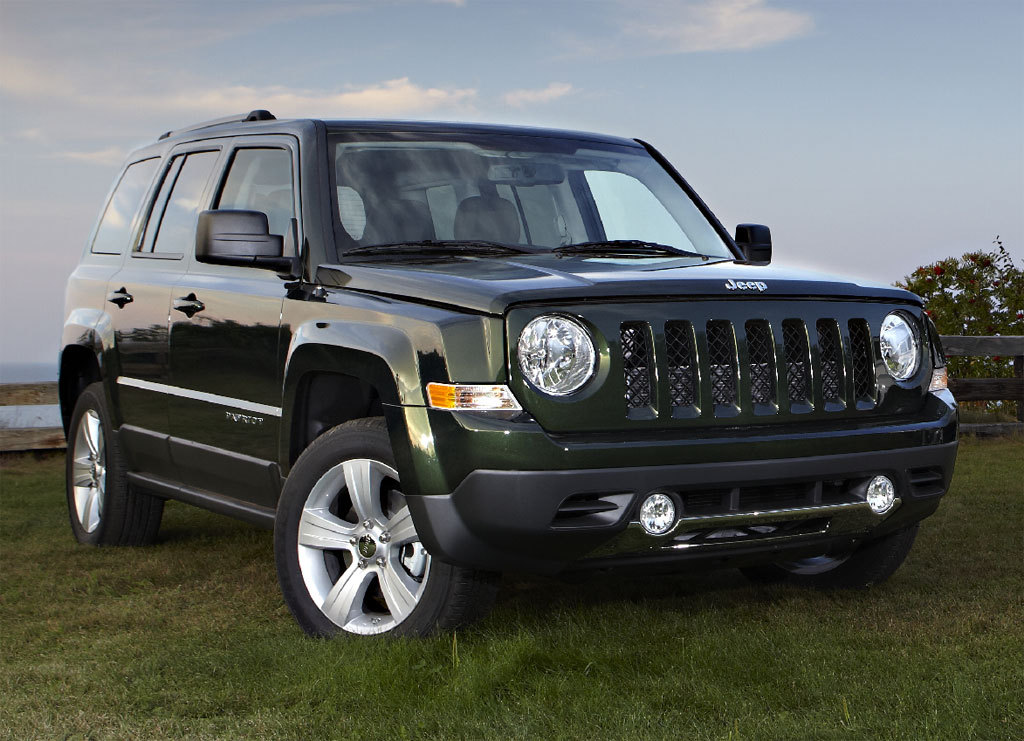}
       \begin{center}
           (a) Clean
       \end{center}
    \end{minipage}
    \hfill
    \begin{minipage}{0.11\textwidth}
        \centering
       \includegraphics[width=\linewidth]{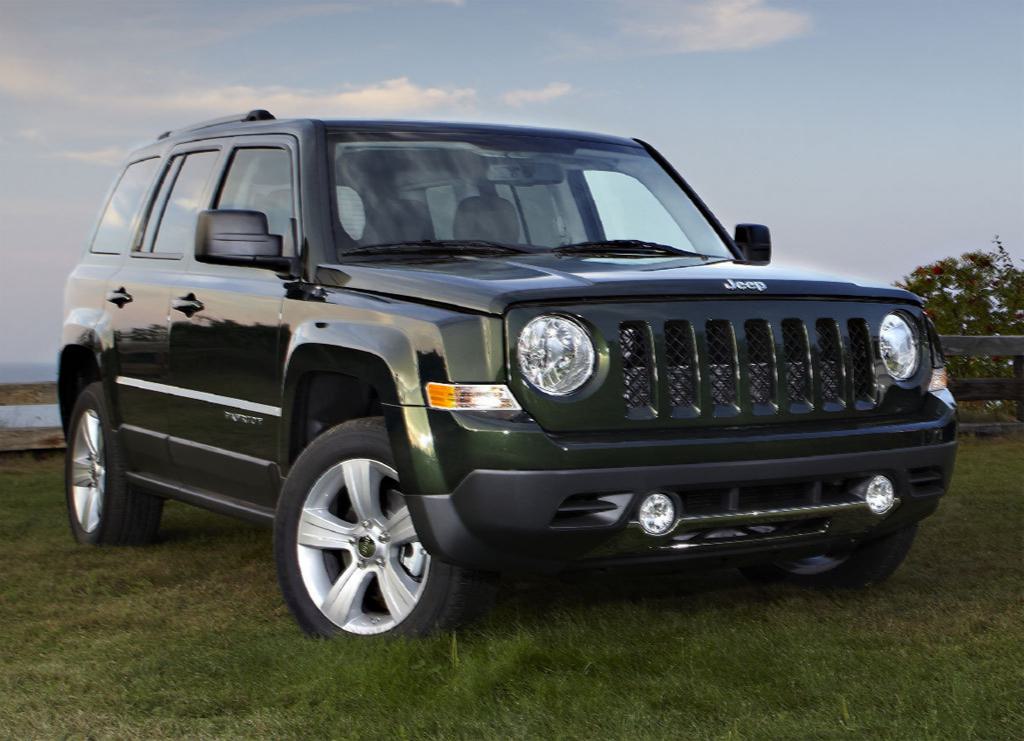}
       \begin{center}
           (b) Motion
       \end{center}
    \end{minipage}
    \hfill
    \begin{minipage}{0.11\textwidth}
        \centering
        \includegraphics[width=\linewidth]{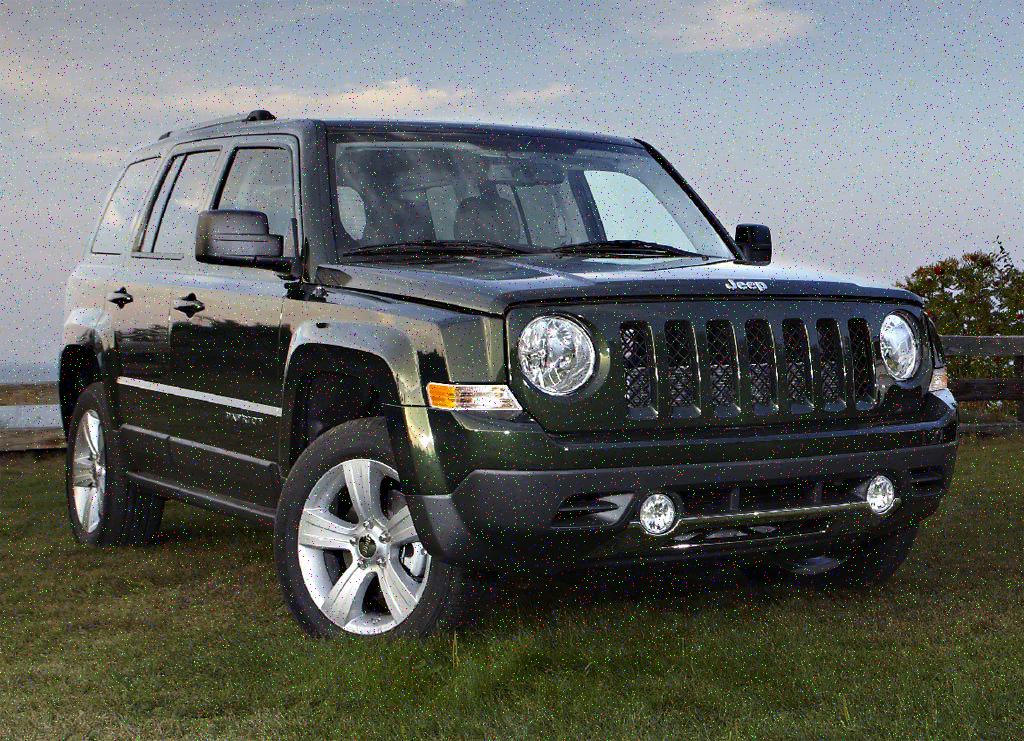}
       \begin{center}
           (c) Impulse
       \end{center}
    \end{minipage}
    \hfill
    \begin{minipage}{0.11\textwidth}
        \centering
       \includegraphics[width=\linewidth]{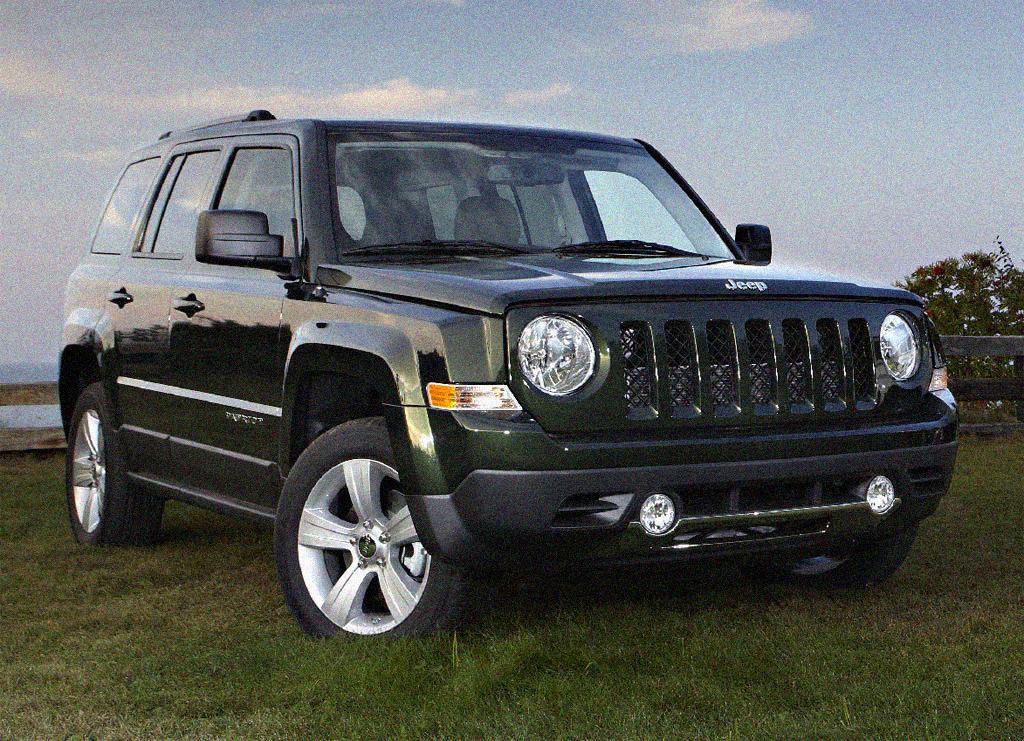}
       \begin{center}
           (d) Gaussian
       \end{center}
    \end{minipage}
    
    \vspace{0.5cm} 
    
    \begin{minipage}{0.11\textwidth}
        \centering
       \includegraphics[width=1\linewidth, height=0.7\linewidth]{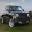}
       \begin{center}
           (e) Pixelate
       \end{center}
    \end{minipage}
    \hfill
    \begin{minipage}{0.11\textwidth}
        \centering
       \includegraphics[width=\linewidth]{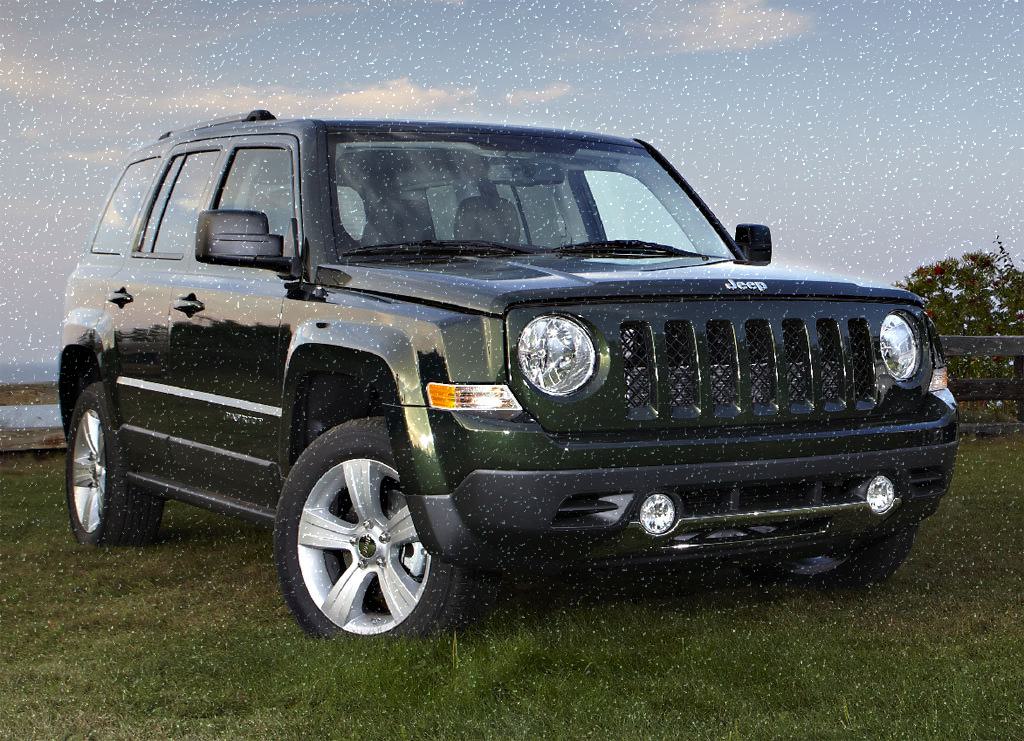}
       \begin{center}
           (f) Spatter
       \end{center}
    \end{minipage}
    \hfill
    \begin{minipage}{0.11\textwidth}
        \centering
        \includegraphics[width=\linewidth]{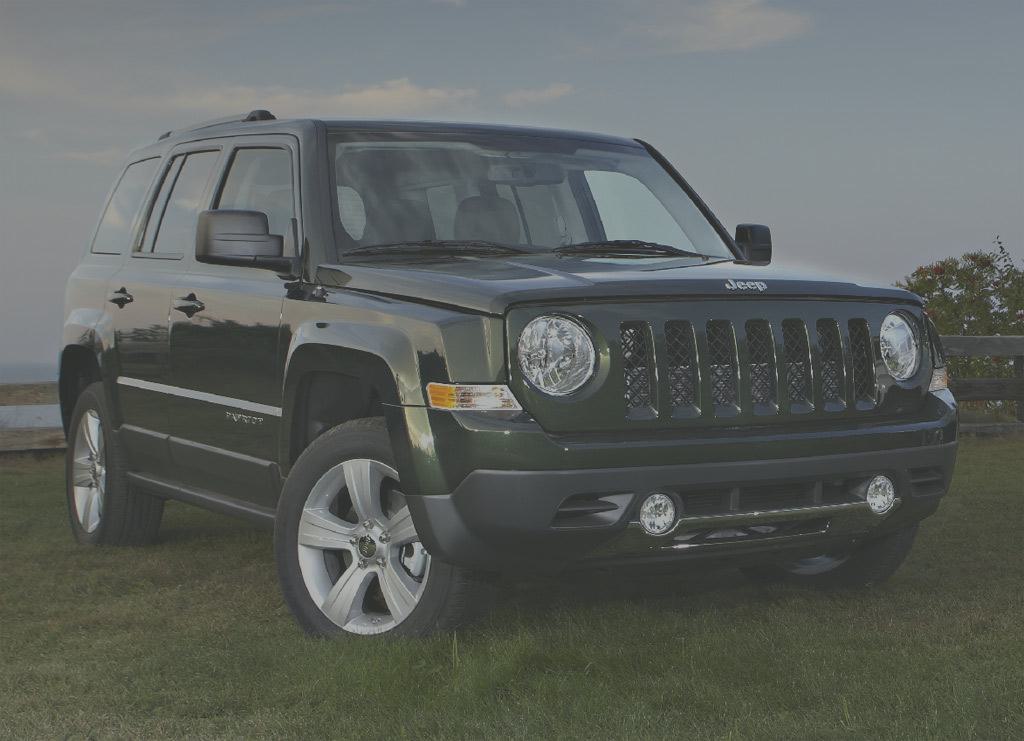}
       \begin{center}
           (g) Contrast
       \end{center}
    \end{minipage}
    \hfill
    \begin{minipage}{0.11\textwidth}
        \centering
       \includegraphics[width=\linewidth]{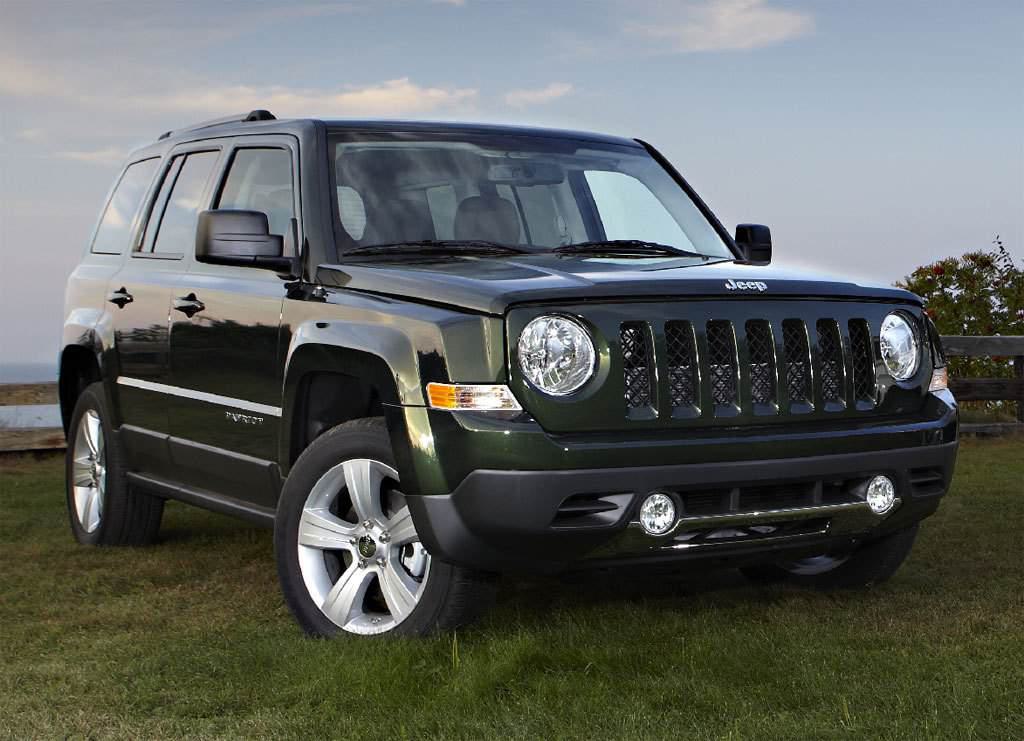}
       \begin{center}
           (h) JPEG
       \end{center}
    \end{minipage}
    \caption{
      Here are eight instances of distorted images, produced using the method suggested in~\citep{hendrycksBenchmarkingNeuralNetwork2019}.
    }
    \label{fig:distorted_images}
    \vspace{-10pt}
\end{figure}

In real-world applications, models often encounter out-of-distribution (OOD) data, where the unlabeled test data may differ from the distribution of the training data. To assess the robustness of our approach, we conducted experiments on a clean test dataset as well as seven corrupted test datasets, including Cars \citep{stallkamp_man_2012}, EuroSAT \citep{helber2018introducing}, RESISC45 \citep{cheng_remote_2017}, and GTSRB \citep{stallkamp_man_2012}. Figure \ref{fig:distorted_images} presents eight sample images from the corrupted Cars's data. We compare our WEMoE and E-WEMoE methods with the Task Arithmetic~\citep{ilharcoEditingModelsTask2023}, Ties-Merging~\citep{yadavResolvingInterferenceWhen2023}, and AdaMerging~\citep{yangAdaMergingAdaptiveModel2023}. 

\begin{table*}
  \caption{Ablations of the test data distribution on ViT-B/16 (for all methods, $\lambda=0.3$).}
  \label{table:abalation_data_distribution_vit_b_16}
  \begin{center}
    \fontsize{8}{9}\selectfont
    \begin{sc}
      \begin{tabular}{l|ccccc|ccccc}
        \toprule
        \textbf{Method}          & \textbf{Cars}                                           & \textbf{EuroSAT}                                          & \textbf{RESISC45} & \textbf{GTSRB} & \textbf{Avg.} & \textbf{Cars} & \textbf{EuroSAT} & \textbf{RESISC45} & \textbf{GTSRB} & \textbf{Avg.} \\
        \midrule
                                 & \multicolumn{5}{c|}{{Clean Test Set}}                   & \multicolumn{5}{c}{{Corrupted Test Set (Motion Blur)}}                                                                                                                                                 \\
        Task Arithmetic          & 75.3                                                    & 96.3                                                      & 85.3              & 80.5           & 84.3          & 73.5          & 70.9             & 83.9              & 72.2           & 75.1          \\
        Ties-Merging             & 74.8                                                    & 93.4                                                      & 84.0              & 65.8           & 79.5          & 73.1          & 65.5             & 82.3              & 57.4           & 69.6          \\
        AdaMerging               & {83.4}                                                  & {97.2}                                                    & {88.6}            & {97.5}         & {91.7}        & {81.3}        & 75.9             & {87.4}            & {95.6}         & {85.0}        \\
        \rowcolor{mygray}
        \textbf{E-WEMoE (2-layer)} &85.4 &98.6 &95.6 &99.0 &94.6 &83.9 &\textbf{82.9} &95.1 &98.1 &\textbf{90.0} \\
        \rowcolor{mygray}
        \textbf{WEMoE (2-layer)} & \textbf{87.3}                                           & \textbf{99.3}                                             & \textbf{96.2}     & \textbf{99.3}  & \textbf{95.5} & \textbf{86.3} & {76.8}    & \textbf{95.2}     & \textbf{98.2}  & {89.1} \\
        \midrule
                                 & \multicolumn{5}{c|}{Corrupted Test Set (Impluse Noise)} & \multicolumn{5}{c}{Corrupted Test Set (Gaussian Noise)}                                                                                                                                                \\
        Task Arithmetic          & 70.4                                                    & \textbf{59.5}                                             & 75.2              & 54.0           & 64.8          & 72.2          & \textbf{60.8}    & 78.5              & 51.0           & 65.6          \\
        Ties-Merging             & 70.5                                                    & 46.2                                                      & 73.0              & 42.0           & 57.9          & 72.8          & 47.6             & 77.0              & 42.2           & 59.9          \\
        AdaMerging               & 77.6                                                    & 42.1                                                      & 81.9              & 90.2           & {73.0} & 79.1          & 58.9             & 81.2              & {74.5}  & \textbf{73.4} \\
        \rowcolor{mygray}
        \textbf{E-WEMoE (2-layer)} &80.8 &49.1 &\textbf{92.4} &96.1 &\textbf{79.6} 
        &83.3 &20.9 &93.8 &\textbf{87.4} &71.4 \\
        \rowcolor{mygray}
        \textbf{WEMoE (2-layer)} & \textbf{83.2}                                           & 11.1                                                      & {92.3}     & \textbf{96.2}  & 70.7          & \textbf{84.8} & 11.7             & \textbf{94.4}     & 73.3           & 66.1          \\
        \midrule
                                 & \multicolumn{5}{c|}{Corrupted Test Set (Pixelate)}      & \multicolumn{5}{c}{Corrupted Test Set (Spatter)}                                                                                                                                                       \\
        Task Arithmetic          & 3.8                                                     & 38.0                                                      & 24.8              & 71.3           & 34.5          & 72.1          & 58.4             & 79.9              & 60.1           & 67.6          \\
        Ties-Merging             & \textbf{4.9}                                            & 36.3                                                      & 21.4              & 57.6           & 30.1          & 72.1          & 50.7             & 77.8              & 46.9           & 61.9          \\
        AdaMerging               & 4.1                                                     & \textbf{46.4}                                             & {23.6}            & {91.3}         & \textbf{41.3} & {79.3}        & \textbf{60.9}    & {85.8}            & {93.7}         & {80.0} \\
        \rowcolor{mygray}
        \textbf{E-WEMoE (2-layer)} &0.5 &22.0 &2.3 &96.7 & 30.4
        &82.0 &58.2 &93.6 &97.3 &\textbf{82.8} \\
        \rowcolor{mygray}
        \textbf{WEMoE (2-layer)} & 0.5                                                     & 20.6                                                      & 1.9               & \textbf{97.3}  & 30.1          & \textbf{84.0} & 11.9             & \textbf{93.8}     & \textbf{97.5}  & 71.8          \\
        \midrule
                                 & \multicolumn{5}{c|}{Corrupted Test Set (Contrast)}      & \multicolumn{5}{c}{Corrupted Test Set (JPEG Compression)}                                                                                                                                              \\
        Task Arithmetic          & 73.4                                                    & 62.5                                                      & 81.3              & 76.9           & 73.5          & 75.1          & 73.1             & 84.8              & 64.7           & 74.4          \\
        Ties-Merging             & 73.4                                                    & 58.0                                                      & 80.0              & 63.1           & 68.6          & 74.8          & 66.9             & 83.8              & 54.1           & 69.9          \\
        AdaMerging               & {81.4}                                                  & {68.1}                                             & {85.8}            & 96.8           & 83.0          & {81.9}        & 76.0             & 87.3              & {91.0}         & {84.1}        \\
        \rowcolor{mygray}
        \textbf{E-WEMoE (2-layer)} &84.1 &\textbf{75.6} &94.5 &98.8 &\textbf{88.2} 
        &84.8 &81.1 &95.4 &94.6 &89.0 \\
        \rowcolor{mygray}
        \textbf{WEMoE (2-layer)} & \textbf{86.0}                                           & 67.8                                                      & \textbf{95.1}     & \textbf{98.9}  & \textbf{87.0} & \textbf{86.9} & \textbf{82.0}    & \textbf{96.2}     & \textbf{95.9}  & \textbf{90.2} \\
        \bottomrule
      \end{tabular}
    \end{sc}
  \end{center}
  \vskip -0.1in
\end{table*}

\begin{figure*}[t]
  \begin{center}
    \includegraphics[width=\linewidth]{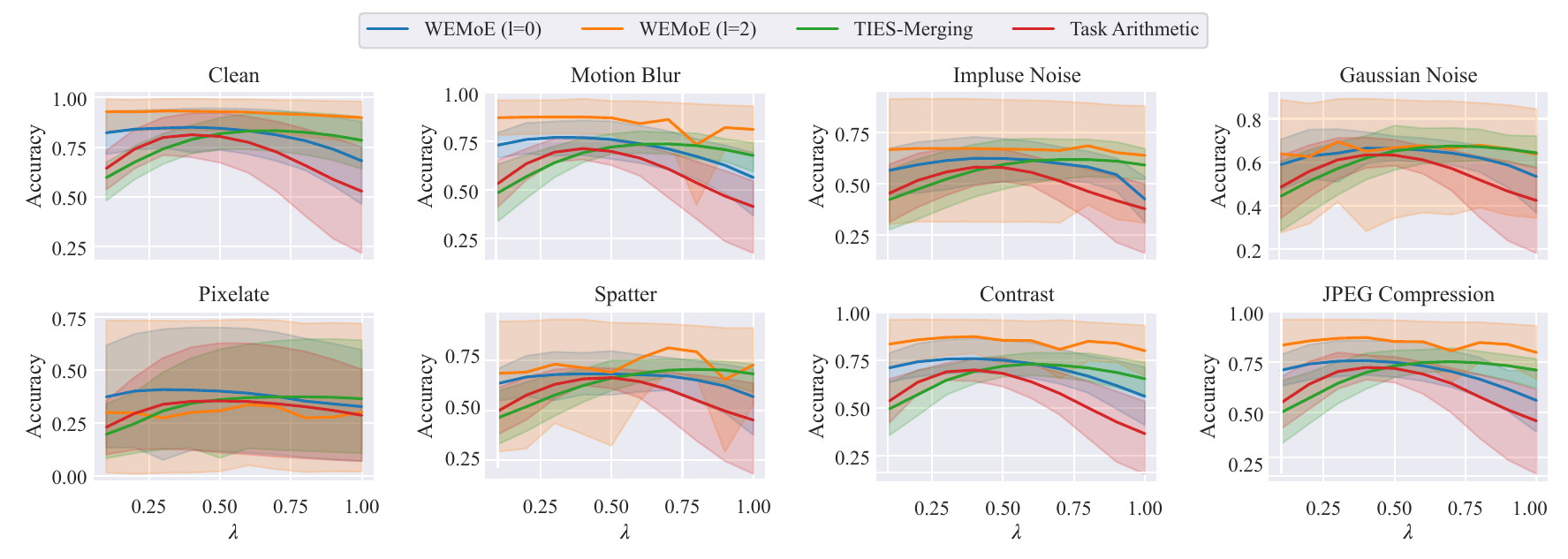}
  \end{center}
  \vskip -0.2in
  \caption{
    The results for robustness experiment on CLIP-ViT-B/32.
    The x-axis of each plot represents the scaling coefficient $\lambda$ of task vectors, while the y-axis shows the accuracy of the merged model on different merged tasks.
  }
  \label{fig:b32_robustness_lambda}
  \vskip -0.15in
\end{figure*}

\begin{figure*}
  \begin{center}
    \includegraphics[width=\linewidth]{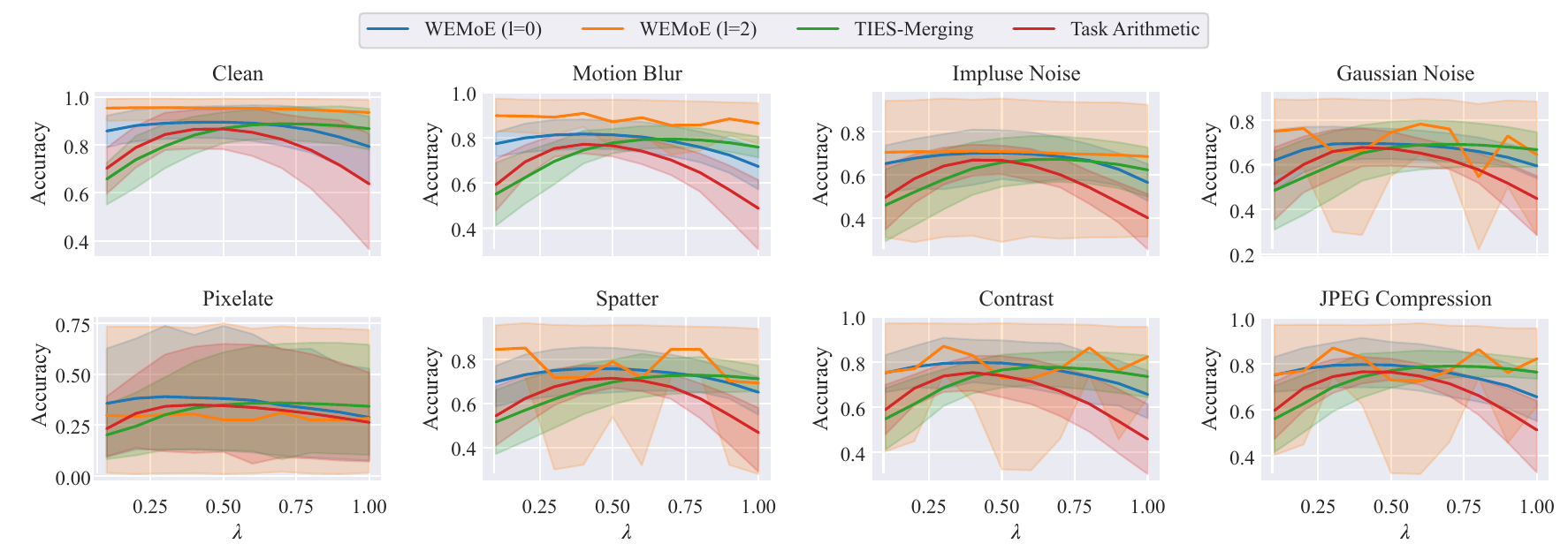}
  \end{center}
    \vskip -0.2in
  \caption{
    The results for the robustness experiment on CLIP-ViT-B/16.
    The x-axis of each plot represents the scaling coefficient $\lambda$ of task vectors, while the y-axis shows the accuracy of the merged model on different merged tasks.
  }
  \label{fig:b16_robustness_lambda}
  \vskip -0.1in
\end{figure*}

The results in Table \ref{table:abalation_data_distribution_vit_b_16} indicate that WEMoE and E-WEMoE consistently achieve the highest performance across both clean and most distorted testsets, highlighting the robustness of the proposed method in handling various data conditions. However, in cases where image quality deteriorates significantly, such as with `Pixelate', our approach may overfit specific tasks, leading to performance drops. Nevertheless, our method remains the most robust in the majority of scenarios.

Figure \ref{fig:b32_robustness_lambda} and Figure \ref{fig:b16_robustness_lambda} show detailed robustness results for CLIP-ViT-B/32 and CLIP-ViT-B/16, respectively, assessing performance under varying merging coefficients. Specifically, the $x$-axis represents the merging coefficient $\lambda$ (in Eq.~\ref{eq:taskarithmetic}) of the task vector, while the $y$-axis indicates the merging model's accuracy across the four tasks. The key observations are as follows: WEMoE ($l=2$) generally outperforms other methods across all tasks and datasets, demonstrating the strongest stability with respect to changes in the hyperparameter $\lambda$. Furthermore, when the test dataset aligns with the training distribution (i.e., `Clean'), WEMoE achieves near-optimal performance for all $\lambda$ configurations.

\section{Analysis}

\subsection{Analysis of the Fine-tuning Configurations}

In this section, we assess the performance variations of architectures trained with different learning rates across various fine-tuning steps.

\begin{figure}[h]
    \centering
    \begin{minipage}{0.24\textwidth}
        \centering
        \includegraphics[width=\linewidth]{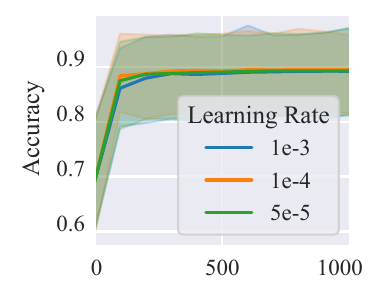}
  \vspace{-20pt}
      \begin{center}
          \text{(a) Learning rate comparison.}
      \end{center}
    \end{minipage}
    \hfill
    \begin{minipage}{0.24\textwidth}
        \centering
         \includegraphics[width=\linewidth]{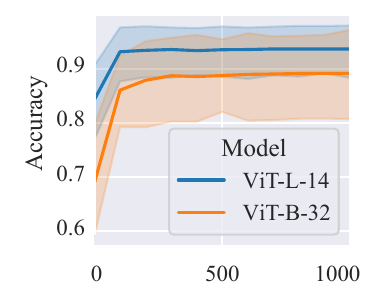}
  \vspace{-20pt}
         \begin{center}
          (b) Model comparison 
      \end{center}
    \end{minipage}
    \caption{
    The performance of the merged models with different numbers of steps:
      (a) CLIP-ViT-B/32 model results using various learning rates.
     (b) A comparison between CLIP-ViT-B/32 and CLIP-ViT-L/14 models.
    }
    \label{fig:wemoe_fusion}
\vskip -0.2in
\end{figure}

Figure \ref{fig:wemoe_fusion} (a) shows the results of merging eight CLIP-ViT-B/32 models, each fine-tuned with different learning rates (e.g., 1e-3, 1e-4, 5e-5). Across all learning rate configurations, the merged model's performance consistently improves as the number of test-time adaptation steps increases, quickly converging to a high accuracy level within just 200 steps. The impact of different learning rates is minimal, indicating that the method is robust to learning rate variations, thus reducing the need for extensive hyperparameter tuning.

Figure \ref{fig:wemoe_fusion} (b) presents the performance trends for the CLIP-ViT-B/32 and CLIP-ViT-L/14 models during test-time adaptation at different steps. The results show that both architectures achieve rapid performance gains as adaptation progresses, reaching optimal levels around 200 steps, further confirming the method's fast convergence across different ViT architectures.

\subsection{Analysis of the Total Number of Parameters}

Table \ref{table:parameter_counts_detailed} compares the total parameter reduction achieved by the E-WEMoE and WEMoE models relative to maintaining independent models for each task when merging different numbers of tasks with the ViT-B/32 architecture. The observations are as follows: (1) As the number of tasks increases, the parameter-saving benefits of WEMoE and E-WEMoE become more pronounced. For instance, when merging 4 tasks, the independent approach requires $\small 4 \times 113.45M=453.8M$ parameters, whereas WEMoE requires a total of $\small 347.25M$ parameters, and E-WEMoE only needs $\small 136.71M$. Additionally, for 8 tasks, the parameter requirements for the independent models, WEMoE, and E-WEMoE are $\small 8 \times 113.45M=907.6M$, $\small 573.96M$, and $\small 159.38M$, respectively. Compared to the independent models, E-WEMoE saves a substantial $\small 748.20M$ parameters, demonstrating its significant efficiency advantage. (2) The cross-layer shared routing strategy in E-WEMoE ensures that the increase in trainable parameters remains minimal as the number of tasks grows, enhancing efficiency.

Table \ref{table:parameter_counts} further illustrates the proportion of trainable parameters relative to the total number of parameters across different architectures and routing layer configurations. We observe that when the number of routing layers is $0$, the model introduces only a few hundred parameters. However, with the $0$ routing layer, the configuration behaves similarly to a special case of AdaMerging~\cite{yangAdaMergingAdaptiveModel2023}, resulting in limited performance gains. With $2$ routing layers, E-WEMoE has significantly fewer trainable parameters compared to WEMoE, achieving reductions of 12, 12, and 24 times on ViT-B/32, ViT-B/16, and ViT-L/14, respectively, thanks to its cross-layer shared routing strategy.

\begin{table*}[t]
  \caption{Comparison of parameter counts and reductions in WEMoE models with varying tasks.}
  \label{table:parameter_counts_detailed}
  \centering
  \small
  \begin{tabular}{lccc}
    \toprule
    \textbf{Method}          & \textbf{Trainable Parameters} & \textbf{Total Parameters} & \textbf{Parameters Reduced by Merging} \\
    \midrule
    Single Pre-trained       & 113.45M (100\%)               & 113.45M                   & -                                      \\
     \midrule
    \rowcolor{mygray}
    WEMoE (2-layer, 2 tasks) & 7.11M (3.04\%)                & 233.89M                   & -6.99M                                 \\
    \rowcolor{mygray}
    WEMoE (2-layer, 3 tasks) & 7.11M (2.45\%)                & 290.57M                   & 49.78M                                 \\
    \rowcolor{mygray}
    WEMoE (2-layer, 4 tasks) & 7.12M (2.02\%)                & 347.25M                   & 106.55M                                \\
    \rowcolor{mygray}
    WEMoE (2-layer, 5 tasks) & 7.13M (1.77\%)                & 403.93M                   & 163.32M                                \\
    \rowcolor{mygray}
    WEMoE (2-layer, 6 tasks) & 7.14M (1.55\%)                & 460.61M                   & 220.09M                                \\
    \rowcolor{mygray}
    WEMoE (2-layer, 7 tasks) & 7.15M (1.38\%)                & 517.28M                   & 276.87M                                \\
    \rowcolor{mygray}
    WEMoE (2-layer, 8 tasks) & 7.16M (1.25\%)                & 573.96M                   & 333.64M                                \\
     \midrule
     \rowcolor{mygray}
    E-WEMoE (2-layer, 2 tasks) &     0.59M (0.47\%)             &  125.37M                   &    101.51M                           \\
    \rowcolor{mygray}
    E-WEMoE (2-layer, 3 tasks) &    0.59M (0.45\%)               & 131.04M                   &    209.29M                            \\
    \rowcolor{mygray}
    E-WEMoE (2-layer, 4 tasks) &   0.59M    (0.43\%)             &   136.71M                  &   317.08M                             \\
    \rowcolor{mygray}
    E-WEMoE (2-layer, 5 tasks) &    0.59M  (0.41\%)              & 142.38M                   &    424.86M                            \\
    \rowcolor{mygray}
    E-WEMoE (2-layer, 6 tasks) &    0.59M   (0.39\%)             & 148.04M                  &     532.64M                           \\
    \rowcolor{mygray}
    E-WEMoE (2-layer, 7 tasks) &   0.59M   (0.38\%)              & 153.71M                    &    640.42M                            \\
    \rowcolor{mygray}
    E-WEMoE (2-layer, 8 tasks) &   0.59M    (0.37\%)             & 159.38M                    &  748.20M                              \\
    \bottomrule
  \end{tabular}
\end{table*}

\begin{table}[t]
  \caption{Parameter count of the up-scaled models from eight tasks.}
  \label{table:parameter_counts}
  \vskip -0.15in
  \begin{center}
    \begin{small}
      \begin{sc}
        \begin{tabular}{lccc}
          \toprule
          \textbf{Model} & \textbf{Trainable} & \textbf{Total} & \textbf{Ratio} \\
          \midrule
          \multicolumn{4}{c}{WEMOE ($l=0$)}                                             \\
          CLIP-ViT-B/32  & 96                 & 566.80M        & 0.00\%         \\
          CLIP-VIT-B/16  & 96                 & 565.15M        & 0.00\%         \\
          CLIP-VIT-L/14  & 192                & 1.95B          & 0.00\%         \\
          \midrule
          \multicolumn{4}{c}{WEMOE ($l=2$)}                                             \\
          CLIP-ViT-B/32  & 7.16M              & 573.96M        & 1.25\%         \\
          CLIP-VIT-B/16  & 7.16M              & 572.31M        & 1.25\%         \\
          CLIP-VIT-L/14  & 25.39M             & 1.98B          & 1.28\%         \\
           \midrule
           \multicolumn{4}{c}{E-WEMOE ($l=2$)}      \\
          CLIP-ViT-B/32  &      0.59M       & 159.38M        &  0.37\%         \\
          CLIP-VIT-B/16  &      0.59M       & 157.70M        &  0.37\%         \\
          CLIP-VIT-L/14  &      1.05M       & 504.70M          &  0.20\%         \\
          \bottomrule
        \end{tabular}
      \end{sc}
    \end{small}
  \end{center}
\vskip -0.2in
\end{table}

\subsection{Analysis of Router Depth}
\label{appendix:ablations_of_router_depth}

To investigate the effect of router depth on the performance of the WEMoE and E-WEMoE models, we conducted an ablation analysis on the router depth. Before delving into the analysis, let's revisit the concept of router depth in our WEMoE model, as described in Section \ref{subsection:weight_ensembling_moe_module}.
In our WEMoE module, the router is implemented as multiple fully connected layers. When the number of routing layers is set to $\{1, 2\}$, the formal representation is as follows:
\begin{equation*}
\small
  \texttt{R}_l(h) \text{ or } \texttt{R}_{shared}(h) = 
  \left\{
  \begin{array}{l}
    \mathbf{W}_1 \texttt{ReLU}(\mathbf{W}_0 h + \mathbf{b}_0) + \mathbf{b}_1, \; \text{if} \; l_{fc}=2, \\
     \mathbf{W}_0 h + \mathbf{b}_0, \; \text{if} \; l_{fc}=1. \\
  \end{array}
  \right.
\label{eq:routing_depth}
\end{equation*}

Tables \ref{table:router_depth_parameter_count} and \ref{table:router_depth} show a comparison of trainable parameter counts and performance between E-WEMoE (1-layer), E-WEMoE (2-layer), WEMoE (1-layer), and WEMoE (2-layer). The results indicate that WEMoE (1-layer) has 73.8K trainable parameters, representing only 0.01\% of the total parameters of the merged model, whereas E-WEMoE (1-layer) is even more efficient, with just 6.15K trainable parameters. Notably, both E-WEMoE (1-layer) and WEMoE (1-layer) outperform AdaMerging and achieve performance that closely matches E-WEMoE (2-layer) and WEMoE (2-layer) across all tasks. This highlights the effectiveness of WEMoE, as well as the efficiency and effectiveness of our E-WEMoE.

\begin{table}[h]
  \caption{
    Parameter comparison of WEMoE (1-layer) and WEMoE (2-layer) on CLIP-ViT-B/32 models.
    We add AdaMerging as a baseline for comparison.
  }
  \label{table:router_depth_parameter_count}
  \centering
  \small
  \begin{tabular}{lc}
    \toprule
    \textbf{Method}         & \textbf{Number of Trainable Parameters} \\
    \midrule
    AdaMerging (layer-wise) & 1.3K                                    \\
     \midrule
    \rowcolor{mygray}
    WEMoE (1-layer)         & 73.8K(0.01\%)                           \\
    \rowcolor{mygray}
    WEMoE (2-layer)         & 7.16M(1.25\%)                           \\
    \midrule
     \rowcolor{mygray}
    E-WEMoE (1-layer)         &      6.15K(0.00\%)                       \\
    \rowcolor{mygray}
    E-WEMoE (2-layer)          &      0.59M(0.37\%)                      \\
    \bottomrule
  \end{tabular}
\end{table}

\begin{table*}[h]
  \caption{
    Ablation study of the router depth on the performance of the up-scaled CLIP-ViT-B/32 models.
    We add AdaMerging as a baseline for comparison.
  }
  \label{table:router_depth}
  \begin{center}
    \setlength{\tabcolsep}{3pt}
    \begin{small}
      \begin{sc}
        \begin{tabular}{lccccccccc}
          \toprule
          \textbf{Method}         & \textbf{SUN397} & \textbf{Cars} & \textbf{RESISC45} & \textbf{EuroSAT} & \textbf{SVHN} & \textbf{GRSRB} & \textbf{MNIST} & \textbf{DTD} & \textbf{Avg.} \\
          \midrule
          AdaMerging (layer-wise) & 66.6            & 68.3          & 82.4              & 92.5             & 86.5          & 93.7           & 97.7           & 61.1         & 80.9          \\
          \midrule
          \rowcolor{mygray}
          E-WEMoE (1-layer)         &73.7             & 75.8           &92.6               &96.9               & 95.6          & 97.8          &    99.3         &   76.9       &    88.6       \\
          \rowcolor{mygray}
          E-WEMoE (2-layer)    & \textbf{74.3}   & 76.3  &  92.7   &  97.9   &  96.1 & 98.6 &   99.5 & \textbf{77.8}    & 89.1 \\
          \midrule
          \rowcolor{mygray}
          WEMoE (1-layer)         & 73.2            & 76.7          & \textbf{93.8}              & 98.6             & 95.7          & 98.6           & 99.5           & 74.5         & 88.3          \\
          \rowcolor{mygray}
          WEMoE (2-layer)         & 74.1            & \textbf{77.4}          & 93.7              & \textbf{99.1}             & \textbf{96.2}          & \textbf{98.9}           & \textbf{99.6}           & 76.4         & \textbf{89.4}          \\
          \bottomrule
        \end{tabular}
      \end{sc}
    \end{small}
  \end{center}
\end{table*}

\begin{figure*}[t]
  \centering
  \includegraphics[width=0.49\linewidth]{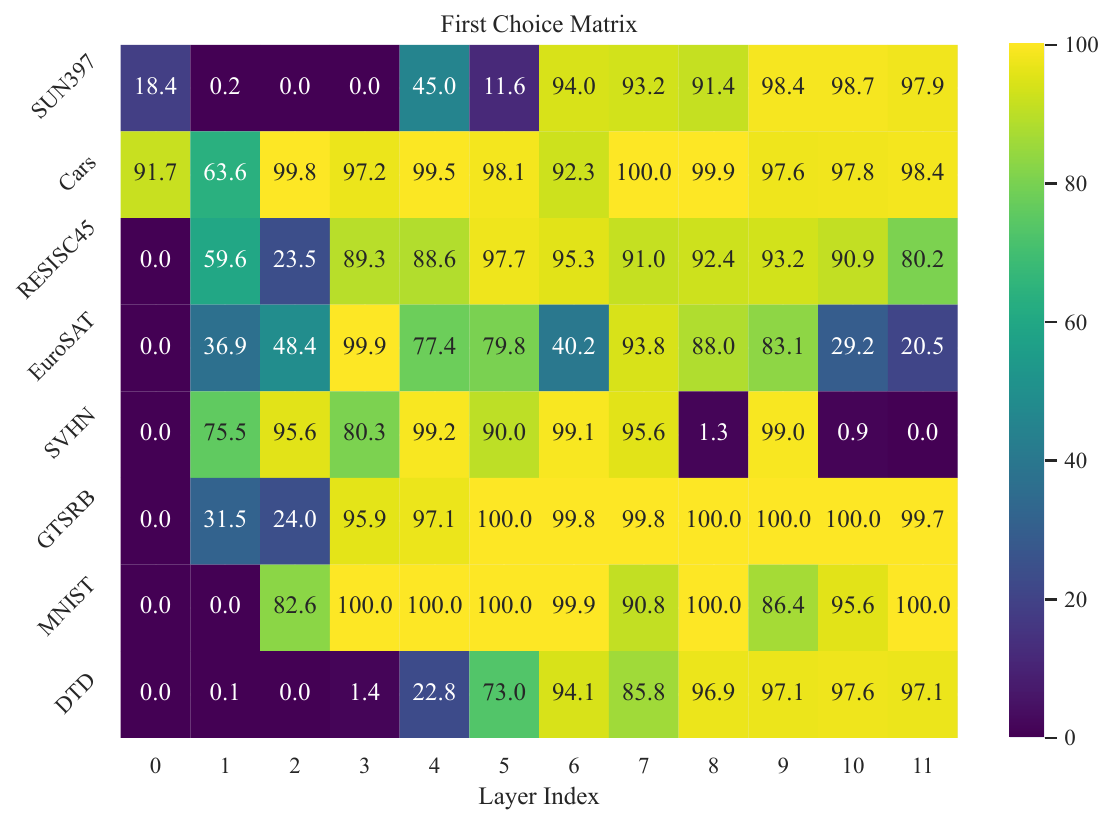}
   \includegraphics[width=0.49\linewidth]{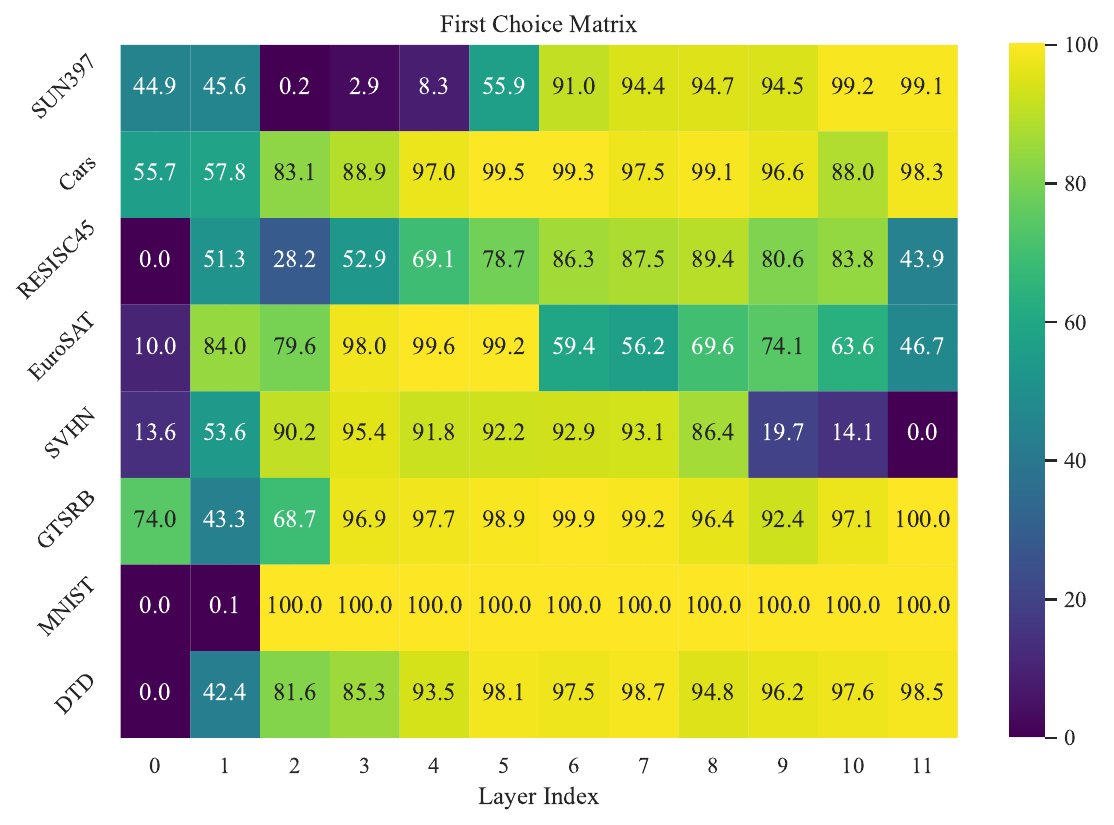}
  \caption{
   Left: First choice matrix of the CLIP-ViT-B/32 model with respect to \textbf{WEMoE}. Right: First choice matrix of the CLIP-ViT-B/32 model with respect to \textbf{E-WEMoE-90\%}. Each row corresponds to a specific task, and the $x$-axis label indicates the layer index. Each entry in the matrix represents the percentage of samples for which the router assigns the highest weight to the corresponding task.
  }
  \label{fig:first_choice_matrix}
\end{figure*}

\subsection{Analysis of Routing Output}
\label{appendix:routing_analysis}

\begin{figure*}[t]
  \centering
  \includegraphics[width=0.95\linewidth]{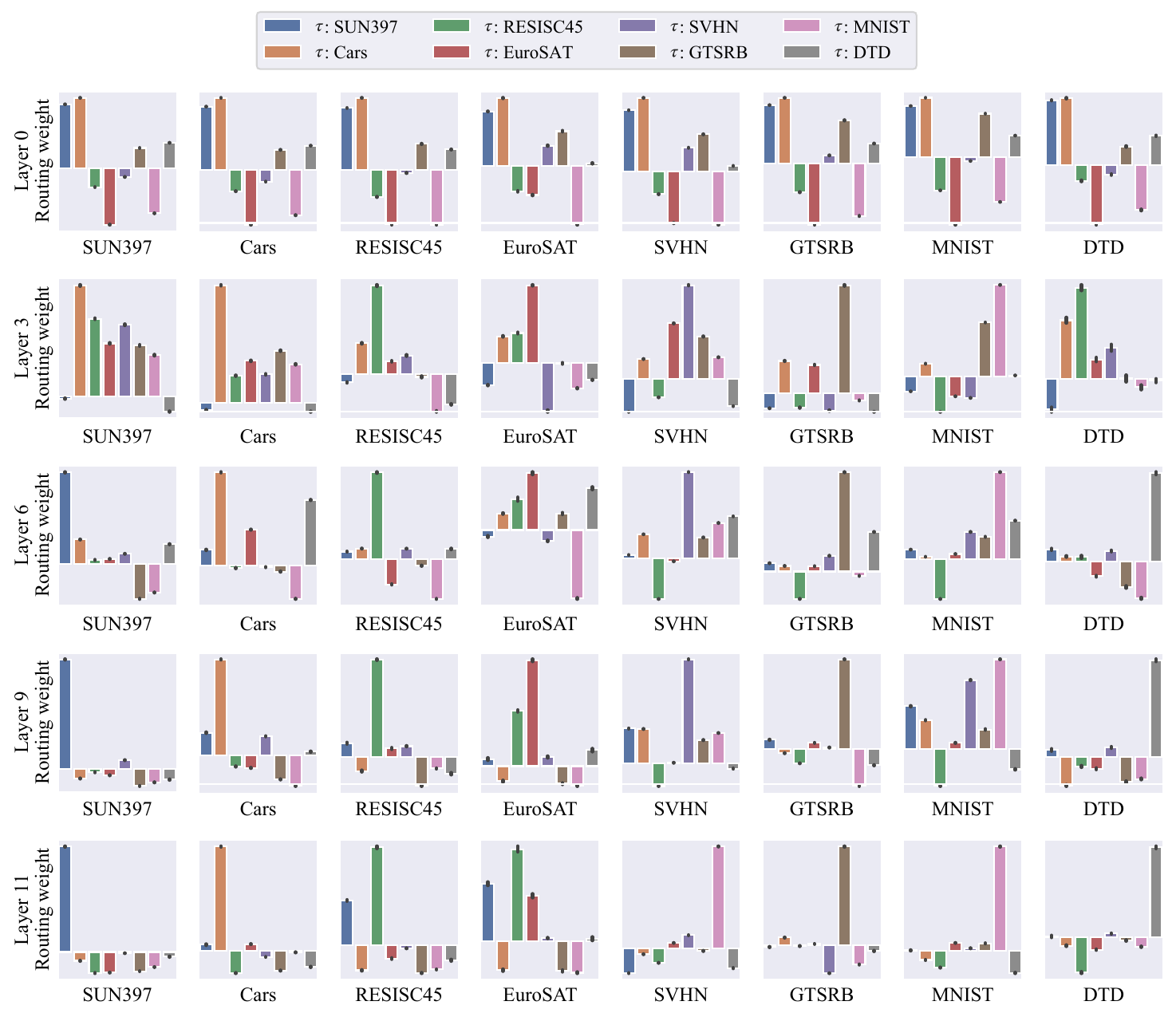}
  \caption{
  Weight distribution analysis of the layer-wise routers in the \textbf{WEMoE} for the CLIP-ViT-B/32 at layers 0, 3, 6, 9, and 11. This figure illustrates the routing weights for various layers and tasks within the neural network. Each subgraph corresponds to a specific task, with the y-axis representing the routing weight of that task and the $x$-axis label indicating the task name.
  }
  \label{fig:router_analysis}
\end{figure*}

\begin{figure*}[t]
  \centering
  \includegraphics[width=0.95\linewidth]{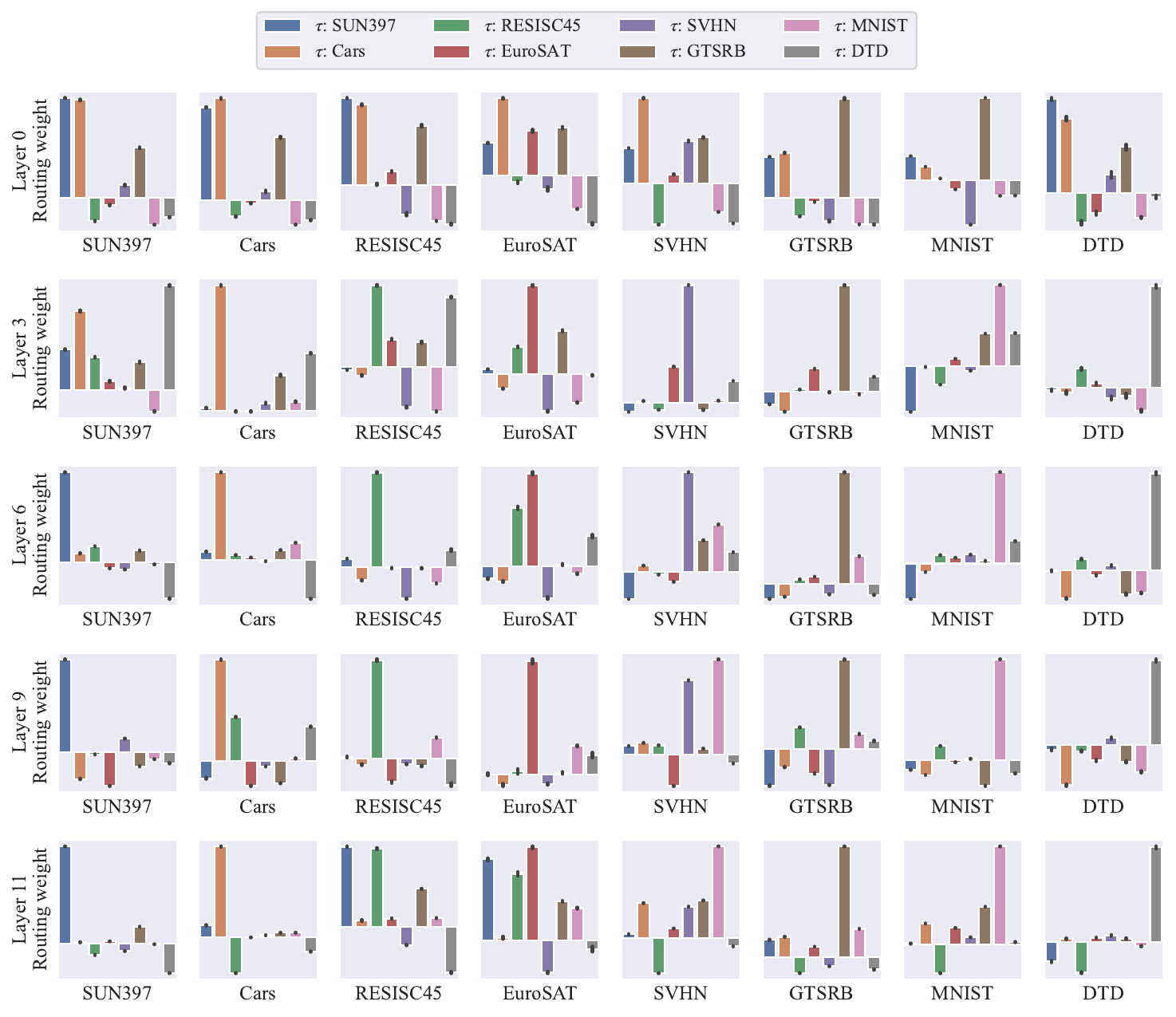}
  \caption{
  Weight distribution analysis of the shared router in the \textbf{E-WEMoE-90\%} for the CLIP-ViT-B/32 at layers 0, 3, 6, 9, and 11. This figure illustrates the routing weights for various layers and tasks within the neural network. Each subgraph corresponds to a specific task, with the y-axis representing the routing weight of that task and the $x$-axis label indicating the task name.
  }
  \label{fig:ewemoe_router_analysis}
\end{figure*}

In this section, we discuss a detailed analysis of the MLP expert weights determined by the routers. This analysis utilizes the CLIP-ViT-B/32 model merged with eight downstream tasks. The results for WEMoE and E-WEMoE are depicted in Figure \ref{fig:router_analysis} and \ref{fig:ewemoe_router_analysis}. Each figure illustrates how routers at varying depths in the network assign routing weights to inputs from different tasks. This visualization enhances our understanding of how routing weights fluctuate across tasks and layers, offering valuable insights into the network's decision-making process.

We observe that for shallow routers, routing weight assignments are not significantly influenced by the source tasks of the input samples, resulting in similar routing weights for samples from different tasks. This phenomenon occurs because the underlying parameters demonstrate greater cross-task versatility. In contrast, deep routers exhibit clear correlations between weight assignments and the source tasks of input samples. In these instances, the router tends to favor the corresponding task vector of the input sample by assigning a higher weight to its source task. This observation suggests that deeper routers are more adept at identifying experts who may perform better based on the input features.

Figure \ref{fig:first_choice_matrix} illustrates the first choice matrix of the CLIP-ViT-B/32 model, representing the percentage of samples to which the router assigns the highest weight for the corresponding task. In the WEMoE model, the routers in the first layer predominantly assign all samples to the MLP expert vector associated with CARS. As the layer index increases, the routers progressively allocate more samples to the correct tasks. By the sixth layer, the routers assign the highest weight to the correct task for most samples. This finding aligns with our previous observation in Figure \ref{fig:router_analysis}, where we noted that the routers in the sixth layer first exhibit a clear correlation between the routing weight assignments and the source task of the input samples. In E-WEMoE, the routers in the first layer allocate most samples to the MLP expert vector corresponding to GTSRB. For deeper routing, our E-WEMoE demonstrates behavior consistent with that of our WEMoE method.

\subsection{Analysis of the Magnitude of MLP Layer}

\begin{figure*}[t]
  \begin{center}
  \centerline{\includegraphics[width=0.9\linewidth]{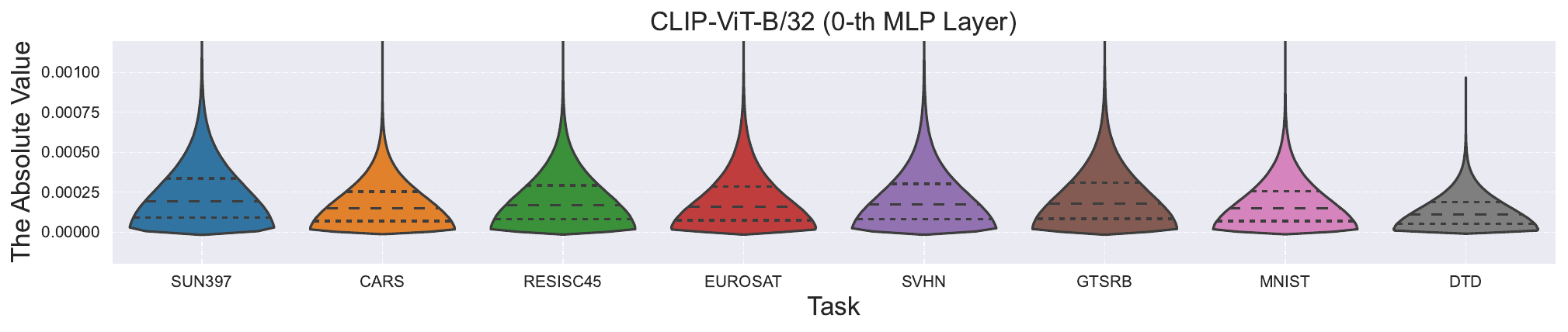}}
  \centerline{\includegraphics[width=0.9\linewidth]{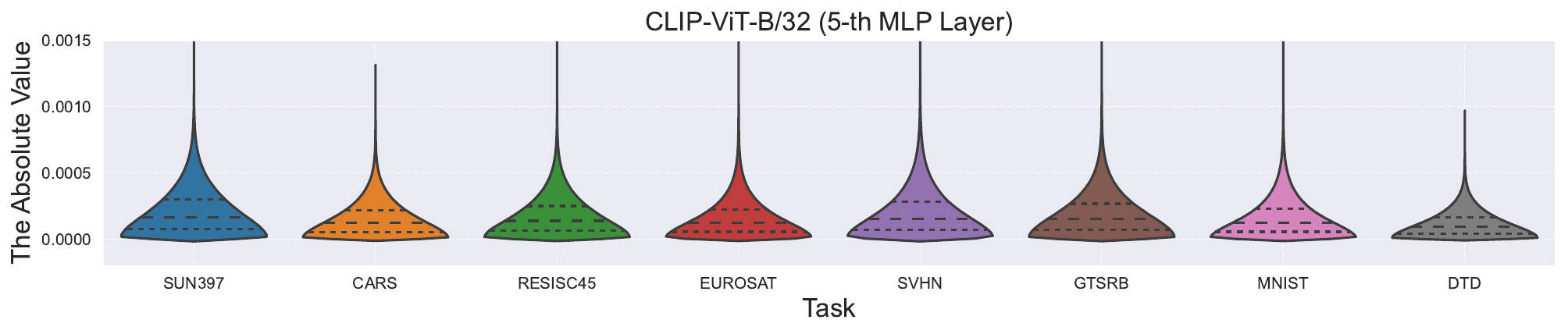}}
   \centerline{\includegraphics[width=0.9\linewidth]{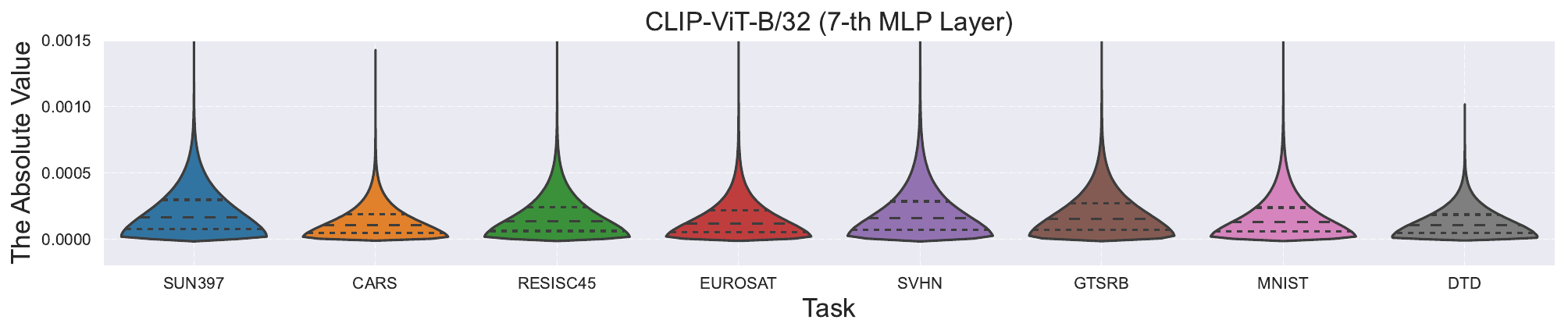}}
    \centerline{\includegraphics[width=0.9\linewidth]{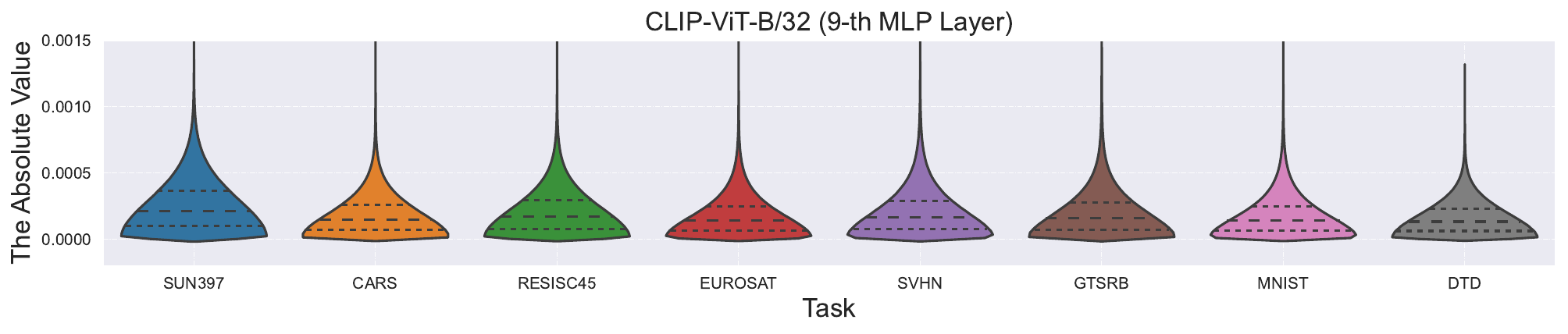}}
     \centerline{\includegraphics[width=0.9\linewidth]{figure/violinplot_vitb32_layer_11_ylim.pdf}}
    \vskip -0.1in
 \caption{
    An illustration of the element magnitudes in the task vector constructed by the MLP in the $l$-th ($l \in \{0,5,7,9,11\}$) Transformer block of ViT-B/32. 
    }
   \label{fig:violinplot_vitb32_appendix}
  \end{center}
  \vskip -0.2in
\end{figure*}

In Section \ref{subsec:ewemoe}, we noted that one of the primary motivations for the E-WEMoE method arises from the observation that in the task vectors constructed by the MLP, most elements exhibit very small magnitudes (in absolute value). To validate this motivation thoroughly, we employ violin plots to visualize the magnitude distribution of the MLP task vectors across different layers (0, 5, 7, 9, and 11) in the ViT-B/32 architecture. As illustrated in Figure \ref{fig:violinplot_vitb32_appendix}, we observe the following: (1) In all layers, the MLP task vectors demonstrate small magnitudes; specifically, the absolute values of over 75\% of the elements across all layers are less than 0.0005. (2) As we progress from shallow to deep layers, the magnitude of the elements in the MLP task vector exhibits a decreasing trend. This trend can be attributed to the fact that the lower layers are more likely to be shared across tasks, necessitating a larger number of elements to accommodate common knowledge. In contrast, the upper layers represent the characteristics of specific tasks, requiring only a smaller number of elements. This insight further motivates us to design varying levels of sparsity for different layers in the subsequent steps.

\stopcontents[sections]

\end{document}